\documentclass[10pt,twocolumn,letterpaper]{article}

\usepackage{booktabs}      % \toprule, \midrule, \bottomrule
\usepackage{siunitx}       % columnas S alineadas en el decimal
\usepackage{threeparttable}% \begin{threeparttable} … \end{threeparttable}

\usepackage[pagenumbers]{cvpr} % To force page numbers, e.g. for an arXiv version

\usepackage{tikz}
\usepackage[abs]{overpic}
\usepackage[normalem]{ulem}

\usepackage[symbol]{footmisc}

\usetikzlibrary{fadings,patterns}

\usepackage{blindtext}
\usepackage{colortbl}
\usepackage{graphicx}
\usepackage{subcaption}
\usepackage{arydshln}
\usepackage[acronym]{glossaries}
\usepackage{bm}
\usepackage{multirow} 
\usepackage{booktabs}

\newcommand{\best}{\cellcolor{tablered}}
\newcommand{\sbest}{\cellcolor{orange}}
\newcommand{\tbest}{\cellcolor{yellow}}

\definecolor{yellow}{rgb}{1, 1, 0.7}
\definecolor{orange}{rgb}{1, 0.85, 0.7}
\definecolor{tablered}{rgb}{1, 0.7, 0.7}
\definecolor{red}{rgb}{1, 0, 0}

\definecolor{cvprblue}{rgb}{0.21,0.49,0.74}
\usepackage[pagebackref,breaklinks,colorlinks,citecolor=cvprblue]{hyperref}

\def\paperID{64} % *** Enter the Paper ID here
\def\confName{3DV\xspace}
\def\confYear{2026\xspace}

\title{GALA: Guided Attention with Language Alignment for Open Vocabulary Gaussian Splatting}
\author{
    Elena Alegret$^{1,2,6}\thanks{Equal contribution.}$ \quad
    Kunyi Li$^{1,4}\footnotemark[1]$ \quad
    Sen Wang$^{1,4}$\quad
    Siyun Liang$^{5}$\\
    Michael Niemeyer$^{3}$ \quad
    Stefano Gasperini$^{1,4,7}$ \quad
    Nassir Navab$^{1,4}$ \quad 
    Federico Tombari$^{1,3}$ \vspace{0.4em} \\
    {\normalsize $^1$Technical University of Munich} \quad 
    {\normalsize $^2$Universitat Politècnica de Catalunya} \quad
    {\normalsize $^3$Google} \\
    {\normalsize $^4$Munich Center for Machine Learning}  \quad 
    {\normalsize $^5$University of Tubingen} \quad
    {\normalsize $^6$ETH Zurich} \quad
    {\normalsize $^7$Visualais} \\
}

\usepackage{graphicx}
\usepackage{etoolbox}

\begin{document}

\vspace{-10mm}
\twocolumn[{%
    \renewcommand\twocolumn[1][]{#1}%
    \maketitle
    \thispagestyle{empty}
    \begin{center}
        \includegraphics[width=1.0\linewidth]{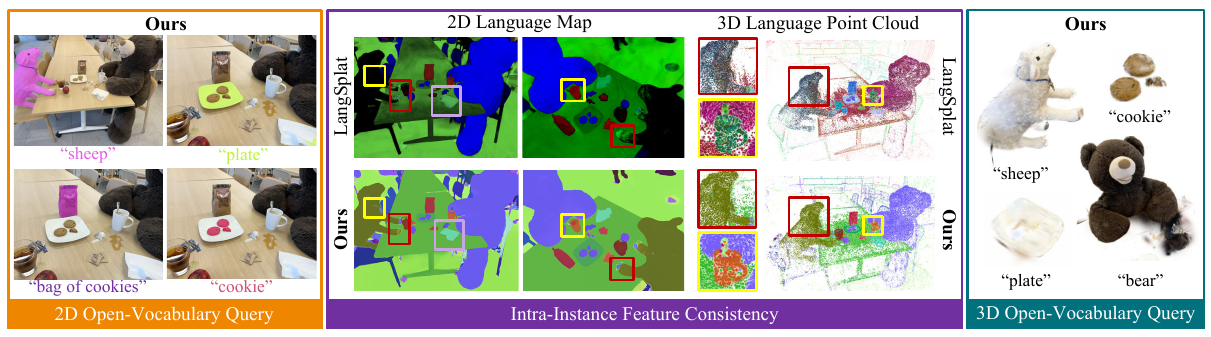}
        \vspace{-5mm}

        \captionof{figure}{We present GALA, a novel 3DGS–based framework for open-vocabulary scene understanding. It delivers strong performance in both 2D and 3D open-vocabulary queries, while preserving high intra-instance feature consistency to boost segmentation quality.}
        \label{fig:teaser}
    \end{center}
}]

\maketitle
\footnotetext[1]{Equal contribution.}
\footnotetext[2]{This work was conducted during Elena's exchange program at TUM.}
% \nolinenumbers
\begin{abstract}
\label{sec:abstract}
3D scene reconstruction and understanding have gained increasing popularity, yet existing methods still struggle to capture fine-grained, language-aware 3D representations from 2D images. In this paper, we present GALA, a novel framework for open-vocabulary 3D scene understanding with 3D Gaussian Splatting (3DGS). GALA distills a scene-specific 3D instance feature field via self-supervised contrastive learning. To extend to generalized language feature fields, we introduce the core contribution of GALA, a cross-attention module with two learnable codebooks that encode view-independent semantic embeddings. This design not only ensures intra-instance feature similarity but also supports seamless 2D and 3D open-vocabulary queries. It reduces memory consumption by avoiding per-Gaussian high-dimensional feature learning. Extensive experiments on real-world datasets demonstrate GALA's remarkable open-vocabulary performance on both 2D and 3D.
\end{abstract}    
\section{Introduction}
\label{sec:intro}

Understanding 3D scenes is a central challenge for 3D computer vision (3DV), with wide-ranging applications in autonomous driving~\cite{geiger2012kitti, caesar2020nuscenes, wang2024drivedreamer, wang2024driving}, robotics~\cite{anderson2018vision, li2024dns, maggio2024clio, hughes2024foundations}, and augmented or virtual reality~\cite{sceneAR2023, rauschnabel2022xr}. Open-vocabulary scene understanding not only enables robotics to perceive and reason about the world but also opens new possibilities for intuitive human-robotic interaction, allowing users to explore and query scenes through natural language. This integration of spatial understanding with language grounding represents a promising direction towards more intelligent systems.

Neural Radiance Fields (NeRF)~\cite{mildenhall2021nerf, barron2021mip, xu2022point, muller2022instant} offer the potential to store additional semantic information within the field. Several methods~\cite{kerr2023lerf, kim2024garfield, ying2024omniseg3d} extend NeRF by distilling semantic or language features from 2D images. However, NeRF-based methods suffer from inefficient encoding and high computational costs for training and rendering. 3D Gaussian Splatting (3DGS)~\cite{kerbl20233d, lu2024scaffold, yu2024mip, huang20242d, niemeyer2025radsplat, li2024geogaussian} provides an explicit and more efficient alternative with a set of 3D Gaussian shape primitives. Subsequent works~\cite{qin2024langsplat, zhou2024feature, liang2024supergseg, jun2025dr, qu2024goi} incorporate feature attributes into these Gaussians, enabling semantic feature rasterization and language-based interactions.

Robotic systems with limited computing resources, for example navigation, is often performed in 2D, even though the robots operate in a 3D world. Therefore, 2D perception remains essential. Recent works~\cite{qin2024langsplat, zhou2024feature} address this by distilling high-dimensional 2D language features~\cite{kirillov2023segment, radford2021learning} into 3D Gaussians through compressing high-dimensional language into low-dimensional representations, followed by novel view synthesis to enable arbitrary-view 2D open-vocabulary querying. However, such compression inevitably leads to information loss, and their segmentation results exhibit low intra-instance consistency and blurred object boundaries, which hinder accurate semantic segmentation.

Instead of prioritizing efficient 2D open-vocabulary segmentation, some works focus on enhancing 3D scene semantics. However, storing high-dimensional language features for each Gaussian is time- and memory-intensive. Recent methods~\cite{wu2024opengaussian, li2025instancegaussian} address this with clustering: Gaussians are grouped into clusters, each assigned a low-dimensional scene-specific feature, and matched to preprocessed per-instance language features via 2D–3D associations. Yet, purely KNN-based clustering without explicit supervision can cause one cluster to span multiple instances or split a single instance, leading to misalignment and degraded segmentation performance. Others~\cite{jun2025dr, cheng2024occam} average multi-view language features without training, achieving strong 3D reconstruction but offering limited or memory-heavy 2D semantic rendering, making them unsuitable for real-time robotics and navigation.

Although reconstructing and perceiving the world in 3D is important and interacting with it in 2D is often the most efficient strategy for robotics~\cite{huang2022visual, anderson2018vision, lei2025gaussnav}, existing approaches tend to focus only on one side of the problem.
% 3D methods~\cite{wu2024opengaussian, li2025instancegaussian, jun2025dr, cheng2024occam} prioritize high-quality 3D semantic reconstruction but often overlook efficiency and provide limited support for accurate 2D segmentation. Conversely, 2D methods~\cite{qin2024langsplat, zhou2024feature} excel at 2D segmentation but, lacking any 3D supervision, perform poorly in 3D semantic understanding and suffer from intra-instance feature inconsistencies, where the semantic features of the same object vary across different spatial positions and viewpoints. 
We propose a \uline{G}uided \uline{A}ttention method with \uline{L}anguage \uline{A}lignment Gaussian Splatting (GALA), a novel framework that enables both 2D and 3D open-vocabulary scene understanding demonstrating the broad applicability to diverse perception tasks, as illustrated in Figure~\ref{fig:teaser}. The key idea is to enforce instance-consistent semantics: instead of storing noisy or redundant per-Gaussian language features, GALA learns to associate each Gaussian with a shared instance-level language embedding, ensuring that the semantics of one instance remain consistent not only across different spatial locations and viewpoints but also in 3D.
Our main contributions can be summarized as follows:
\begin{itemize}
    \item We propose to store per-instance semantics via codebooks, associating each instance with a language embedding and ultimately generating intra-instance consistent semantic features for better segmentation.
    \item By employing an attention mechanism that maps each Gaussian feature to its corresponding instance, we enable effective 2D and 3D open-vocabulary segmentation.
    \item We improve the segmentation with an attention-weighted entropy loss, which encourages a clear one-to-one mapping between Gaussian instance features and codebook embeddings.
\end{itemize}
Extensive experiments on public real-world datasets, LERF-OVS \cite{kerr2023lerf} and ScanNet-v2 \cite{dai2017scannet}, demonstrate the effectiveness of GALA on both 2D and 3D semantic segmentation and open-vocabulary localization compared to the state-of-the-art. The code and models will be released upon acceptance.
\section{Related Works}
\label{sec:related}

\subsection{Zero-Shot 2D Scene Understanding}
The success of 2D visual foundation models has been demonstrated across a wide range of vision tasks, which enhances both perceptual and reasoning abilities.
CLIP~\cite{radford2021learning} aligns image and text features through contrastive learning, enabling robust cross-modal understanding in a shared embedding space. DINO~\cite{oquab2024dinov2}, a self-supervised Vision Transformer, learns rich semantic representations from unlabeled images, capturing object boundaries and scene layouts. Building on these models, Grounding DINO~\cite{liu2024grounding} extends DINO with open-vocabulary detection capabilities guided by textual queries, through tight visual-language fusion.
SAM~\cite{kirillov2023segment}, a promptable segmentation model, enables zero-shot instance segmentation with impressive generalization. Grounded SAM~\cite{ren2024grounded} combines SAM with Grounding DINO to support arbitrary text-driven semantic segmentation and detection. APE~\cite{shen2024aligning} introduces a unified visual perception framework for tasks like segmentation and grounding, using lightweight visual-language fusion for efficient and generalizable performance.
However, these powerful models are inherently limited to 2D image understanding, restricting their applicability in tasks requiring holistic 3D scene understanding.

\subsection{Open-Vocabulary 3D Scene Understanding}
Understanding 3D scenes requires consistent semantic reasoning across multiple views and spatial dimensions. Recent efforts have explored transferring powerful language features from 2D models into 3D representations to allow robots to perceive the world like humans.
OpenScene~\cite{peng2023openscene} distills CLIP features into 3D point clouds for zero-shot segmentation and language queries, but suffers from limited spatial resolution and reduced generalization due to point-based representation.
More recent methods~\cite{kerr2023lerf, kim2024garfield, ying2024omniseg3d, cen2023segment} integrate semantics into continuous neural radiance field by distilling 2D language features, enabling open-vocabulary 3D understanding. However, NeRFs remain slow to render, depend heavily on high-quality 2D masks, and struggle with scalability due to volumetric computation.

In contrast, 3D Gaussian Splatting (3DGS) provides an explicit and efficient representation better suited for real-time 3D understanding. LangSplat~\cite{qin2024langsplat} applies hierarchical feature distillation by assigning each Gaussian a low-dimensional feature that is rasterized into a 2D feature map. A pretrained autoencoder is used to compress high-dimensional language features for supervision. Similarly, Feature3DGS~\cite{zhou2024feature} leverages a convolutional neural network (CNN) for feature dimension lifting. While both methods reduce the dimensionality of the supervision signal, this compression inevitably causes information loss. Furthermore, they learn per-Gaussian semantic features without enforcing intra-instance feature consistency, which may lead to ambiguous object representations and hinder robotic interaction and navigation.
OpenGaussian~\cite{wu2024opengaussian} and InstanceGaussian~\cite{li2025instancegaussian} place greater emphasis on 3D awareness by enabling point-level 3D segmentation through hierarchical feature clustering and 3D–2D feature association, mapping scene-specific instance features to language features. However, misalignment in this mapping can cause significant performance drops.

Rather than training a semantic feature field per scene, Dr.\ Splat~\cite{jun2025dr} and Occam's LGS~\cite{cheng2024occam} propose an aggregation method that averages multi-view language features in a single forward pass, greatly improving efficiency. Although these methods improve 3D semantic reconstruction, generating accurate 2D semantic maps remains crucial for robotics, enabling fast and reliable perception from onboard camera images.
SuperGSeg~\cite{liang2024supergseg} clusters thousands of Gaussians into SuperGaussians sharing language embeddings, enabling efficient high-dimensional feature rendering and improving performance. However, its MLP-based cluster update is complex and may lack semantic coherence, sometimes grouping irrelevant or noisy points. Moreover, the K-Nearest Neighbors (KNN)-based initialization depends on point density, so sparse regions can cause a SuperGaussian to span multiple objects with conflicting semantics, degrading segmentation quality.
GOI~\cite{qu2024goi} and CCL-LGS~\cite{tian2025ccl} both introduce a single trainable feature codebook to store language embeddings and use a multi-layer perceptron (MLP) to predict discrete codebook indices for the rasterized 2D feature maps. While this approach compresses semantics spatially rather than dimensionally preserving semantic richness, the MLP applies fixed weights uniformly across all input elements, lacking the flexibility to dynamically prioritize important information. This limitation makes it less effective at capturing context-dependent relevance compared to attention mechanisms. 

Therefore, we propose a dual-codebook design combined with a guided cross-attention module. Our method computes similarity scores for soft, continuous assignments between Gaussian features and codebook embeddings, enabling instance-level semantics in a differentiable manner. Despite relying on 2D supervision, the fully linear attention and rasterization modules enhance generalization from 2D tasks to 3D tasks and effectively reduce the multi-view inconsistencies found in prior work.

\section{Preliminaries}
\label{sec:preliminary}

3D Gaussian Splatting (3DGS)~\cite{kerbl20233d} employs a set of 3D points to effectively render images from given viewpoints, each characterized by a Gaussian function with 3D mean $\mathbf{\mu}_i \in \mathbb{R}^3$, covariance matrix $\Sigma_i \in \mathbb{R}^{3\times3}$, opacity value $\alpha_i \in \mathbb{R}$, RGB color value $\mathbf{c}_i \in \mathbb{R}^3$, and sometimes with feature value $\mathbf{m}_i \in \mathbb{R}^d$: 
\begin{equation}
    \sigma_i(\mathbf{x}) = \alpha_i * \exp\left( -\frac{1}{2} (\mathbf{x} - \mathbf{\mu}_i)^T \Sigma_i^{-1} (\mathbf{x} - \mathbf{\mu}_i) \right).
\label{eq:Gaussian}
\end{equation}

Given a 3D position $\mathbf{x}$, $\sigma_i(\mathbf{x})$ represents current opacity value contributed by the $i$-th Gaussian. To facilitate optimization, $\Sigma_i = R_iS_i S^T_i R^T_i$ is factorized into the product of a scaling matrix $S_i$, represented by scale factors $\mathbf{s}_i \in \mathbb{R}^3$, and a rotation matrix $R_i$ encoded by a quaternion $\mathbf{r}_i \in \mathbb{R}^4$.
Color value $\mathbf{\hat{C}}(\mathbf{u})$ and feature value ${\mathbf{\hat{M}}}(\mathbf{u})$ at pixel $\mathbf{u}$ are rendered by $N$ projected and ordered Gaussians using point-based $\alpha$-blending:
\begin{equation}
    \{\mathbf{\hat{C}}, \mathbf{\hat{M}} \}(\mathbf{u}) = \sum_{i \in N} T_i \sigma_i \times \{\mathbf{c}_i, \mathbf{m}_i \},
\label{eq:blending}
\end{equation}
where $T_i = \prod_{j=1}^{i-1} (1 - \sigma_j)$. 
Scaffold-GS~\cite{lu2024scaffold} introduces a neural variant of 3DGS by voxelizing a set of point clouds as anchor points $ V \in \mathbb{R}^{N \times 3}$. Each anchor point $ \mathbf{v}_i \in V$ is associated with a feature $ \mathbf{f}_i \in \mathbb{R}^{d}$, scaling factor $l_i \in \mathbb{R}^{3}$ and $K$ learnable offsets $\{\mathcal{O}_{i,k} \in \mathbb{R}^3 \mid k=0, \dots, K-1\}$. Then $K$ neural Gaussians $\{\mu_{i, 0}, \ldots, \mu_{i, K-1}\} = \mathbf{v}_i + \{\mathcal{O}_{i, 0}, \ldots, \mathcal{O}_{i, K-1}\} \cdot l_i$ are generated from a given anchor point $\mathbf{x}_v$. The remaining attributes of each Gaussian $\mathbf{g}_i \in \{ \alpha_{i,k}, \mathbf{c}_{i,k}, R_{i,k}, S_{i,k}, \mathbf{m}_{i,k}\} $ are predicted as:
\begin{equation}
\{ \mathbf{g}_{i, 0}, \ldots, \mathbf{g}_{i, K-1} \} =\mathcal{F}_{\mathbf{g}}(\mathbf{f}_i, \mathbf{\delta_{i}}, \vec{\mathbf{d}_{i}}),
\label{eq:sprawn}
\end{equation}
where $\mathbf{\delta}_{i} = \| \mathbf{v}_i - \mathbf{x}_c \|$, $\vec{\mathbf{d}}_{i} = \frac{ \mathbf{v}_i - \mathbf{x}_c }{ \| \mathbf{v}_i - \mathbf{x}_c \| }$, $ \mathbf{x}_c $ is the camera center, and $\mathcal{F}_{\mathbf{g}}$ is corresponding attribute decoder.

\begin{figure*}[t]
    \centering
    \includegraphics[width=0.9\linewidth]{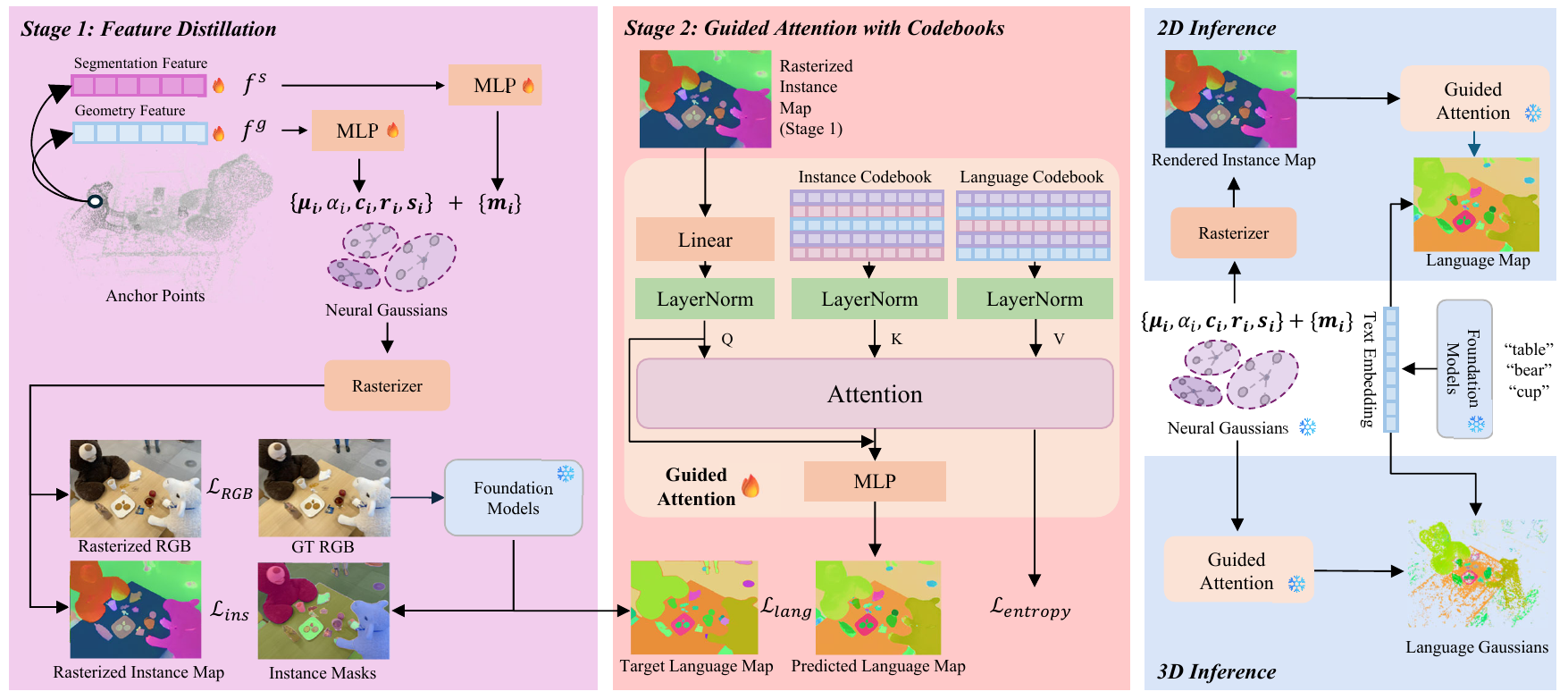}
    \caption{\textbf{Overview of GALA.} In Stage 1, we reconstruct the 3D scene and distill a scene-specific feature field in a self-supervised manner. In Stage 2, a rasterized instance feature map is used to train a Guided Attention module, which learns to map the scene-specific feature field to a generalized language field via two learnable codebooks. During inference (right), GALA supports open-vocabulary querying and segmentation in both 2D (top) and 3D (bottom).}
    \label{fig:gala-overview}
    \vspace{-0.2cm}
\end{figure*}

\section{Methods}
\label{sec:method}
As shown in Figure~\ref{fig:gala-overview}, our method builds on neural Gaussian Splatting~\cite{lu2024scaffold} with two-stage training: (1) self-supervised reconstruction of scene geometry and a scene-specific instance feature field, and (2) rendering these features to 2D and mapping them to generalized language features via guided attention with dual learnable codebooks. The linear attention design enables seamless segmentation in both 2D and 3D using only 2D training, while the per-instance codebooks and attention-weights entropy loss enforce one-to-one mappings, enhancing intra-instance feature consistency.

\subsection{Scene Reconstruction and Feature Distillation}
SuperGSeg~\cite{liang2024supergseg} builds on Scaffold-GS~\cite{lu2024scaffold} to perform joint 3D reconstruction and scene-specific instance and hierarchical feature distillation, where these features are used for clustering. In contrast, our method does not rely on hierarchical features for clustering and instead focuses solely on instance learning in a self-supervised manner.
Consequently, each anchor point in our method is assigned to two types of features. Using Eq.~\ref{eq:sprawn}, a geometry feature $ \mathbf{f}^g_i \in \mathbb{R}^{d_g} $ is decoded into $K$ Gaussian attributes $\{\alpha_{i,k}, \mathbf{c}_{i,k}, R_{i,k}, S_{i,k}\}$, and a segmentation feature $\mathbf{f}^s_i \in \mathbb{R}^{d_{\text{seg}}}$ is decoded into $K$ instance features $\mathbf{m}_{i,k} \in \mathbb{R}^{d_{\text{ins}}}$. 
These attributes and features are then rasterized as a rendered color map $\mathbf{\hat{C}} \in \mathbb{R}^{H \times W \times 3}$ and a instance feature map $\mathbf{\hat{M}} \in \mathbb{R}^{H \times W \times d_{\text{ins}}}$ through Eq.~\ref{eq:blending}.

LangSplat~\cite{qin2024langsplat} and Feature3DGS~\cite{zhou2024feature} learn per-Gaussian semantic features independently, without any instance-level constraints. Our method adopts a self-supervised method \cite{wu2024opengaussian, ying2024omniseg3d} to distill a scene-specific instance field with instance contrastive learning. We first generate a set of instance masks $\{{m}_i \in \mathbb{R} \mid i=0, \dots, \mathcal{M} \}$ for each view using Segment Anything Model (SAM)~\cite{kirillov2023segment}. For a given instance mask ${m}_i$, we denote each pixel feature within the mask as $\{{\hat{m}}_{i, j} \in \mathbb{R}^{d_{\text{ins}}} \mid j = 1, \ldots, n \}$, where subscript $i$ denotes the mask index and subscript $j$ denotes the pixel index. We compute the mean feature value within the mask as $\bar{{m}}_i \in \mathbb{R}^{d_{\text{ins}}}$. To distill the 3D instance field in a self-supervised manner and enhance intra-instance feature similarity, we employ contrastive learning to pull features within the same mask closer together, while pushing features from different masks further apart:
\vspace{-0.2cm}
\begin{equation}
    \mathcal{L}_{\text{ins}} = \frac{1}{\mathcal{M}}\sum_{i=1} \sum_{j=1} - \log \frac{\exp({\hat{m}}_{i,j} \cdot {\bar{m}}_i / \tau_i)}{\sum_{q \neq i}^{\mathcal{M}} \exp({\hat{m}}_{i,j} \cdot {\bar{m}}_q / \tau_q)}.
\end{equation}

Therefore, the overall objective function for the first stage is:
\begin{equation}
\mathcal{L}_{\text{1}} = \mathcal{L}_{\text{RGB}} + \lambda_{\text{ins}} \mathcal{L}_{\text{ins}},
\end{equation}
where $ \lambda_{\text{ins}} $ is the penalty coefficient. And $\mathcal{L}_{\text{RGB}}=0.8 \times |\mathbf{C} - \mathbf{\hat{C}}|+0.2 \times SSIM(\mathbf{C} - \mathbf{\hat{C}})$ is the photometric loss \cite{kerbl20233d} where $\mathbf{C}$ is the ground-truth color image. \\

\subsection{Semantic Codebooks}
\label{sec:codebook}
Prior works such as OpenGaussian~\cite{wu2024opengaussian}, InstanceGaussian~\cite{li2025instancegaussian}, and SuperGSeg~\cite{liang2024supergseg} adopt a bottom-up approach: they first cluster low-level features to form clusters and then learn instance-level segmentation by aggregating these clusters. However, this strategy can lead to several issues, including over-segmentation, one single cluster representing multiple distinct objects, or different parts of the same object being assigned to separate clusters. To address this, we introduce a codebook module designed to represent each instance in the scene with a unique embedding. 
A codebook consists of $N_c$ learnable embeddings, where $N_c$ approximates the number of instances in the scene. Specifically, we define an Instance Codebook $\mathcal{C}_{ins} \in R^{N_c\times d_{ins}}$, where each entry captures a distinct instance-level representation. In parallel, we define a Language Codebook $\mathcal{C}_{lang} \in R^{N_c\times d_{c}}$ which stores language embeddings with a one-to-one correspondence to the entries in $\mathcal{C}_{inst}$. Each codebook entry is intended to represent a unique instance in the scene.
The proposed codebooks decouple semantics from spatial positions and allow for unambiguous, per-instance embedding assignments, ensuring intra-instance feature similarity.

\subsection{Guided Attention with Codebooks}\label{sec:attention-mechanism}
Perceiving the world through human language is a key goal of 3D scene understanding, for which a purely scene-specific feature field is insufficient. Prior works~\cite{wu2024opengaussian, li2025instancegaussian} attempt to align low-dimensional features with high-dimensional language semantics via 2D–3D associations, while others~\cite{qin2024langsplat, zhou2024feature} compress high-dimensional supervision to reduce overhead. However, these approaches are either designed for 3D or 2D segmentation tasks, suffering from information loss, or leading to limited generalization.
Our method introduces a guided cross-attention module with codebooks proposed in Section \ref{sec:codebook} that maps scene-specific features to the generalized language field, enabling both 2D and 3D open-vocabulary queries. 

\noindent \textbf{Attention with Codebooks.}
An attention module \cite{vaswani2017attention} is adopted with learnable codebooks and residual connections. We use the rasterized instance feature map $\mathbf{\hat{M}}$ as the query $Q$, the instance codebook $\mathcal{C}_{ins}$ as the key $K$ and the language codebook $\mathcal{C}_{lang}$ as the value $V$:
\begin{align}
\mathbf{\hat{A}} &= \mathcal{A}(\mathbf{\hat{M}}) = {Attn}(Q, K, V) + Q \in \mathbb{R}^{HW \times d_c},
\label{eq:attn} \\
\mathbf{\hat{L}} &= \mathcal{F}_{lift}(\mathbf{\hat{A}}) \in \mathbb{R}^{HW \times d_{lang}},
\label{eq:lift}
% Q &= \mathcal{N}(M \times W^Q) \in \mathbb{R}^{HW \times d_{ins}}, \\
% K &= \mathcal{N}(C_{\text{ins}}) \in \mathbb{R}^{N_c \times d_{ins}}, \\
% V &= \mathcal{N}(C_{\text{lang}}) \in \mathbb{R}^{N_c \times d_c},
\end{align}
where $Q = \mathcal{N}(\mathbf{\hat{M}} \times W^Q), K = \mathcal{N}(C_{\text{ins}}), V = \mathcal{N}(C_{\text{lang}})$, $\mathcal{N}$ represents layer normalization opertor which is applied to avoid scale discrepancies, $W^Q$ is the linear transformation which is applied to project the original input into a same space of instance codebook and $\mathcal{F}_{lift}$ is an MLP to lift the feature dimensionality.
During training, we apply only 2D supervision with cosine similarity loss between the predicted language map $\mathbf{\hat{L}}$ and the preprocessed ground-truth language map $\mathbf{L}$: 
\begin{equation}
\mathcal{L}_{\text{lang}} = 1 - \cos(\mathbf{\hat{L}}, \mathbf{L}).
\label{eq:cos_loss}
\end{equation}

It is worth noting that the attention operation $\mathcal{A}$ defined in Eq. \ref{eq:attn} is linear. As the rasterization in Eq. \ref{eq:blending} involves a weighted summation, applying $\mathcal{A}$ to the 2D rasterized features is mathematically equivalent to applying it directly to the underlying 3D Gaussians:
\begin{equation}
    \mathcal{A}(\mathbf{\hat{M}}) = \mathcal{A}\big(\sum T_i \sigma_i \times \mathbf{m}_i\big) = \sum T_i \sigma_i \times \mathcal{A}(\mathbf{m}_i).
\end{equation}
This property allows us to train the codebooks solely using 2D feature maps as supervision, and during inference, however, the same model can be directly applied to the 3D Gaussians, enabling open-vocabulary semantic queries in both 2D and 3D space. By compacting per-Gaussian's semantic features into per-instance embeddings, we not only reduce training costs but also enforce intra-instance feature consistency in both 2D and 3D.

\noindent \textbf{Probability Guidance.}
OpenGaussian~\cite{wu2024opengaussian} adopts two-level clustering with positional embedding to model instance-level representations. However, without explicit supervision, it struggles to establish a one-to-one correspondence between instances and clusters. Our method leverages attention weights to guide a clear one-to-one mapping between instances and codebook embeddings. The attention weights:
\begin{equation}
P= softmax({QK^\top}/{\sqrt{d_{ins}}}) \in \mathbb{R}^{HW \times N_c} ,
\end{equation}
indicate the relevance probability of each codebook embedding with respect to each feature query. To encourage a one-to-one correspondence, we apply the entropy loss on the attention weights:
\begin{equation}
\mathcal{L}_{\text{entropy}} = -\sum_{j=1} p_j \log(p_j),
\end{equation}
where $p_j \in P$ represents the attention probability distribution over the $N_c$ codebook entries for feature query $j$. This enforces the probability distribution for each query to be unimodal, meaning each query is associated with a single codebook embedding, ensuring that each instance corresponds to only one embedding in the codebook.

Therefore, the overall objective function of the second stage is:
    \begin{equation}
    \mathcal{L}_{\text{2}} = \mathcal{L}_{\text{lang}} + \lambda_{\text{ent}} \,\mathcal{L}_{\text{entropy}},
    \end{equation}
where $ \lambda_{\text{ent}} $ is the penalty coefficient. \\

\begin{figure*}[t]
    \small
    \centering
    \includegraphics[width=0.85\linewidth]{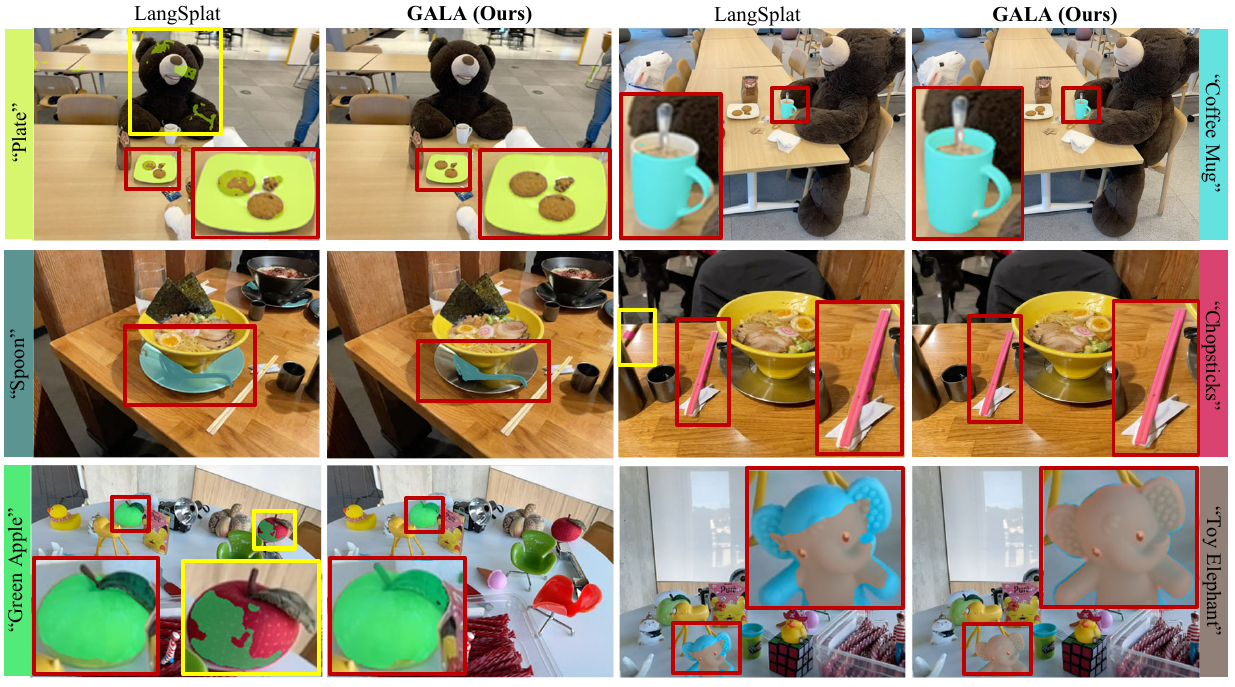}
    \caption{\textbf{Qualitative Results of 2D Open-Vocabulary Query.} We visualize 2D Open-Vocabulary query results on LERF-OVS dataset~\cite{kerr2023lerf}. LangSplat fails to localize objects accurately, leading to mismatching or incomplete masks. Our method delivers precise and consistent queries across diverse queries.
    }
    \label{fig:2d-eval-lerfovs}
\end{figure*}

\begin{table*}[t]
\centering
\small
\resizebox{0.8\textwidth}{!}{
    \begin{tabular}{llccccccccccc}
    \toprule
     & & \multicolumn{2}{c}{\textbf{Mean}} & \multicolumn{2}{c}{\textit{Figurines}} & \multicolumn{2}{c}{\textit{Teatime}} & \multicolumn{2}{c}{\textit{Ramen}} & \multicolumn{2}{c}{\textit{Waldo\_kitchen}} \\
    \cmidrule(lr){3-4} \cmidrule(lr){5-6} \cmidrule(lr){7-8} \cmidrule(lr){9-10} \cmidrule(lr){11-12}
    Eval. & \textbf{Method} & mIoU $\uparrow$ & mAcc $\uparrow$ & mIoU & mAcc & mIoU & mAcc & mIoU & mAcc & mIoU & mAcc \\
    \midrule
    \multirow{7}{*}{\textbf{2D}} 
            & LERF~\cite{kerr2023lerf}            & 37.40 & 73.60 & 38.60 & 75.00 & 45.00 & 84.80 & 28.20 & 62.00 & 37.90 & 72.70  \\
            & LangSplat~\cite{qin2024langsplat}   & \tbest51.40 & \sbest84.30 & 44.70 & \tbest80.40 & \sbest65.10 & \sbest88.10 & \sbest51.20 & \sbest73.20 & \tbest44.50 & \best\textbf{95.50}  \\
            & Feature-3DGS~\cite{zhou2024feature} & 45.70 & 77.00 & 58.80 & 77.20 & 40.50 & 73.40 & 43.70 & \tbest69.80 & 39.60 & 87.60  \\
            & GS-Grouping~\cite{ye2024gaussian}   & 46.30 & 76.50 & \sbest60.90 & 75.00 & 40.00 & 74.30 & 45.50 & 68.60 & 38.70 & 88.20  \\
            & LEGaussians~\cite{shi2024language}  & 46.90 & 77.20 & \tbest60.30 & 75.60 & 40.80 & 75.20 & 46.00 & 67.50 & 39.40 & 90.30  \\
            & GAGS~\cite{peng2024gags} & \sbest54.12 & \tbest81.66 & 53.59 & 78.57 & \tbest60.29 & \best88.14 & \tbest46.81 & 69.01 & \best55.80 & \sbest90.91  \\
            & GOI~\cite{qu2024goi}                & 50.60 & \best\textbf{84.40} & \best\textbf{63.70} & \best\textbf{88.60} & 44.50 & 82.90 & \best\textbf{52.60} & \best\textbf{75.50} & 41.40 & \tbest90.40  \\
            & \textbf{Ours}                       & \best\textbf{55.49}  & 73.43 & 59.35 & \sbest82.14 & \best\textbf{76.73} & \best\textbf{88.14} & 35.13 & 50.70 & \sbest{50.75} & 72.73   \\
    \midrule
    \multirow{5}{*}{\textbf{3D}} 
            & LangSplat~\cite{qin2024langsplat}       & 9.66 & 12.41 & 10.16 & 8.93 & 11.38 & 20.34 & 7.92 & 11.27 & 9.18 & 9.09   \\
            & LEGaussians~\cite{shi2024language}      & 16.21 & 23.82 & 17.99 & 23.22 & 19.27 & 27.12 & 15.79 & \sbest26.76 & 11.78 & 18.18  \\
            & OpenGaussian~\cite{wu2024opengaussian}  & \best\textbf{38.36} & \tbest51.43 & \tbest39.29 & \tbest55.36 & \best\textbf{60.44}  & \tbest76.27 & \best\textbf{31.01} & \best\textbf{42.25} & \tbest22.70 & \tbest31.82  \\
            & SuperGSeg~\cite{liang2024supergseg}     & \tbest35.94 & \sbest52.02 & \sbest43.68 & \sbest60.71 & \sbest55.31  & \sbest77.97 & \sbest18.07 & 23.94 & \sbest26.71 & \sbest45.45  \\
            & \textbf{Ours}                           & \sbest36.71 & \best\textbf{59.71} & \best\textbf{45.25} & \best\textbf{69.64} & \tbest53.27 & \best\textbf{84.75} & \tbest17.08 & \tbest25.35 & \best\textbf{31.22} & \best\textbf{59.09} \\
    \bottomrule
    \end{tabular}
}
\caption{\textbf{2D and 3D Evaluation on LERF-OVS.} We report mIoU and mAcc on the LERF-OVS dataset \cite{kerr2023lerf}. Note that OpenGaussian~\cite{wu2024opengaussian} and SuperGSeg~\cite{liang2024supergseg} by default do not report 2D evaluation. LERF~\cite{kerr2023lerf}, Feature-3DGS~\cite{zhou2024feature}, GS-Grouping~\cite{ye2024gaussian}, GAGS~\cite{peng2024gags} and GOI~\cite{qu2024goi} by default do not support 3D evaluation.}
\label{table:lerf-ovs}
\vspace{-0.2cm}
\end{table*}
% Previous Ramen: 35.13 & 50.70 
% Previous Ramen: \tbest17.08 & \tbest25.35

\section{Experiments}
\label{sec:experiment}
% DONE: Figurines + Ramen + Teatime 19.08 & 2394

\subsection{Experimental Setup}
\noindent \textbf{Datasets.} 
We comprehensively evaluate our method on two real-world datasets: ScanNet-v2 \cite{dai2017scannet} and LERF-OVS \cite{kerr2023lerf}. Following OpenGaussian \cite{wu2024opengaussian}, 8 scenes are selected from the ScanNet-v2.

\noindent \textbf{Baselines.} 
We compare our method in both 2D and 3D with LERF \cite{kerr2023lerf}, LangSplat \cite{qin2024langsplat}, Feature-3DGS~\cite{zhou2024feature}, GS-Grouping~\cite{ye2024gaussian}, LEGaussians \cite{shi2024language}, GOI \cite{qu2024goi}, SuperGSeg~\cite{liang2024supergseg} and OpenGaussian~\cite{wu2024opengaussian}.

\noindent \textbf{Metrics.}
We follow common practice and report open-vocabulary segmentation and object selection evaluation with mean Intersection-over-Union (mIoU) for segmentation accuracy and mean accuracy (mAcc) for localization accuracy. While understanding the world in 3D is essential, perceiving it in 2D offers a more efficient pathway for real-time performance in robotics. Therefore, we report our performance both in 2D and 3D to demonstrate the broad applicability of our method to diverse perception tasks.

\noindent \textbf{Implementation Details.} 
We perform single-GPU training (NVIDIA RTX 3090). For stage 1, we train 30,000 iterations with $\lambda_\text{ins}=0.001$ and for stage 2 we train 15,000 iterations with $\lambda_\text{ent}=10$. We set the dimension of both instance codebook and language codebook as $d_{ins}=d_{c}=16$. We use SAM~\cite{kirillov2023segment} and CLIP~\cite{radford2021learning} to preprocess the ground-truth language map, and set $d_{lang}=512$. For more implementation details, please refer to the supp. mat..

\begin{figure*}[t]
    \small
    \centering
    \includegraphics[width=0.85\linewidth]{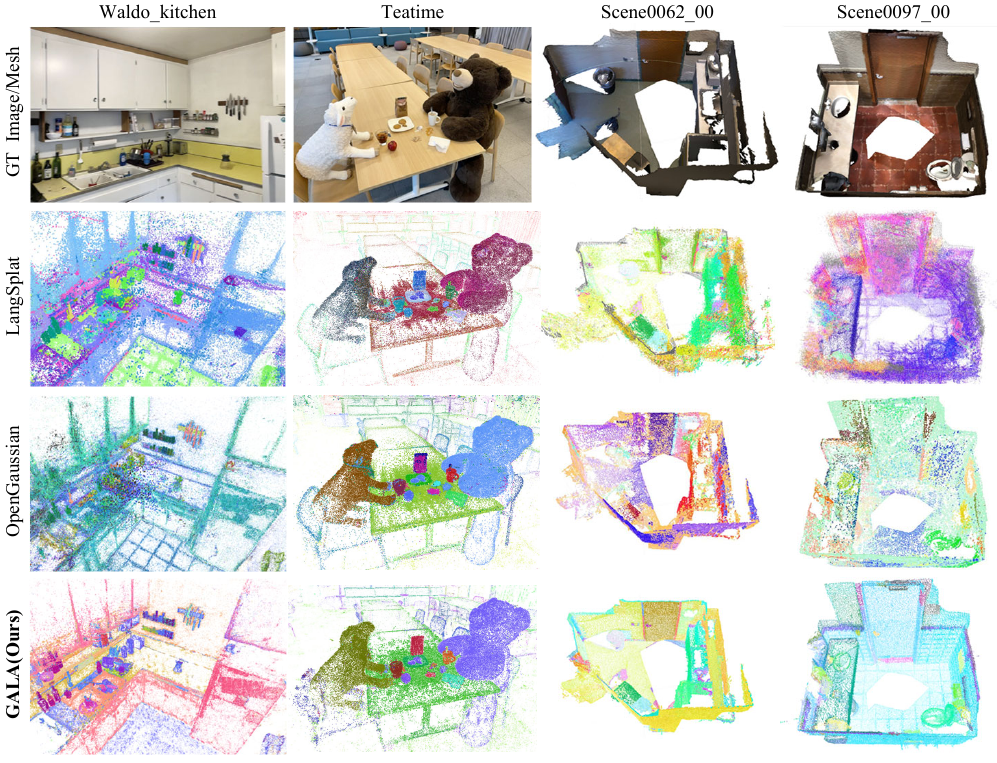}
    \caption{\textbf{Qualitative Results of 3D Open-Vocabulary Segmentation.} We visualize the language feature point cloud on LERF-OVS~\cite{kerr2023lerf} and ScanNet-v2 dataset~\cite{dai2017scannet} by compressing the features into the RGB point cloud. Note that the colors for visualization are consistent only within each method and not method-to-method.}
    \label{fig:3d-eval}
    \vspace{-0.2cm}
\end{figure*}

\subsection{2D Evaluation}
Table~\ref{table:lerf-ovs} presents the 2D results on the LERF-OVS dataset. We report both per-scene and average evaluations, where our method achieves the highest average mIoU (55.49\%) among all existing approaches. In scenes with clearly separated objects, such as \textit{Teatime}, our method delivers precise performance in both open-vocabulary segmentation and localization, achieving 76.73\% mIoU and 88.14\% mAcc. Our method also performs robustly in more cluttered environments like \textit{Waldo\_Kitchen}, attaining 50.75\% mIoU, where it better distinguishes complex domestic objects compared to LangSplat and GOI. Figure~\ref{fig:2d-eval-lerfovs} demonstrates that our method can accurately distinguish the \textit{``Green Apple"} without ambiguity, whereas LangSplat incorrectly selects both the green and red apples. Overall, our method achieves precise object segmentation with sharp and well-defined boundaries.

\subsection{3D Evaluation}
\noindent \textbf{3D Evaluation on LERF-OVS.}
Following the evaluation protocol of LangSplat~\cite{qin2024langsplat}, Table~\ref{table:lerf-ovs} showcases the strong performance of our method in 3D segmentation and localization on the LERF-OVS dataset. Thanks to the linearity of the proposed attention module, our method, trained exclusively with 2D supervision, generalizes seamlessly to 3D tasks without any architectural modifications. Our method even outperforms the 3D-only method OpenGaussian on \textit{Figurines} and \textit{Waldo kitchen}, which, however, cannot easily produce 2D segmentation outputs. In contrast, the 2D-only method LangSplat struggles with 3D evaluation, as it is trained solely with 2D supervision and lacks 3D-aware segmentation. We also visualize the feature point cloud in Figure~\ref{fig:3d-eval}. Our method achieves both better geometry reconstruction and 3D segmentation compared with LangSplat~\cite{qin2024langsplat} and OpenGaussian~\cite{wu2024opengaussian}.

\noindent \textbf{3D Evaluation on ScanNet-v2.}
Table~\ref{table:scannet} reports the 3D point cloud segmentation results on the ScanNet-v2 dataset, as ScanNet-v2 provides ground-truth semantic point cloud. We present the mean mIoU and mAcc across eight selected scenes containing different numbers of classes. Our method consistently outperforms OpenGaussian on all metrics, delivering strong 3D reconstruction and segmentation accuracy alongside high-quality 3D localization. Figure~\ref{fig:3d-eval} visualizes the language-featured point clouds.
By default, OpenGaussian does not densify Gaussians on ScanNet-v2, resulting in sparse features and lower appearance quality. Our method surpasses OpenGaussian both quantitatively and qualitatively.

% Average IoU: 0.1708
% Acc@0.25: 0.2535

\begin{table}[t]
\setlength{\tabcolsep}{3pt}
\centering
\small
\resizebox{\linewidth}{!}{
\begin{tabular}{lcccccc}
\toprule
%\multirow{2}{*}{\textbf{Method}} 
& \multicolumn{2}{c}{\textbf{19 Classes}} 
& \multicolumn{2}{c}{\textbf{15 Classes}} 
& \multicolumn{2}{c}{\textbf{10 Classes}} \\
\cmidrule(lr){2-3} \cmidrule(lr){4-5} \cmidrule(lr){6-7}
\textbf{Method} & mIoU $\uparrow$ & mAcc $\uparrow$ 
& mIoU $\uparrow$ & mAcc $\uparrow$ 
& mIoU $\uparrow$ & mAcc $\uparrow$ \\
\midrule
LanSplat~\cite{qin2024langsplat}        
& 2.94 & 11.63 & 3.80 & 13.98 & 6.60 & 22.24 \\
OpenGaussian~\cite{wu2024opengaussian}  
& 15.47 & 26.04 & 17.42 & 28.82 & 23.46 & 37.73 \\
% OpenGaussian* 
% & \sbest16.98 & \tbest25.93 & \sbest20.03 & \sbest30.64 & \sbest27.46 & \sbest40.71 \\
\textbf{Ours} 
& \textbf{21.54} & \textbf{37.47} & \textbf{25.20} & \textbf{42.06} & \textbf{35.85} & \textbf{57.02} \\
\bottomrule
\end{tabular}
} 
\caption{\textbf{3D Evaluation on ScanNet-v2.} We report the average 3D mIoU and mAcc on 8 scenes of the ScanNet-v2 dataset~\cite{dai2017scannet}.}
\label{table:scannet}
\vspace{-0.4cm}
\end{table} 
% NOTE:
% For best result:        \best\textbf{} 
% For second best result: \sbest
% For third best result:  \tbest

\subsection{Intra-Instance Feature Consistency}
\begin{figure}[t]
    \small
    \centering
    \includegraphics[width=0.75\linewidth]{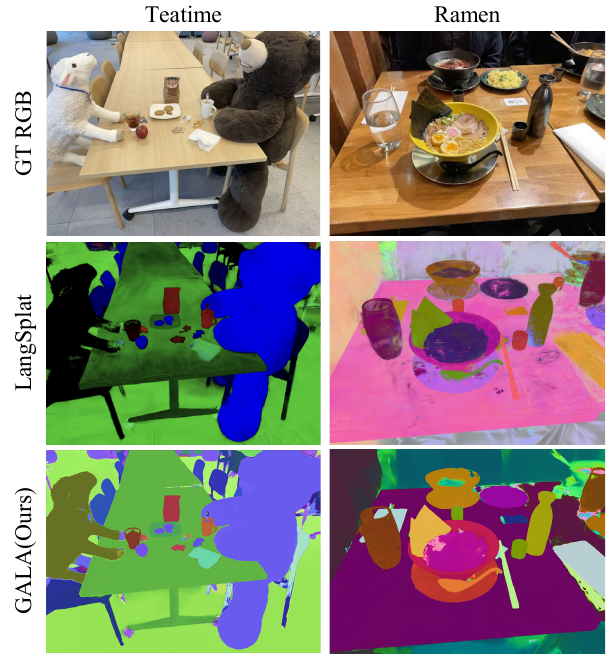}
    \caption{\textbf{Intra-Instance Feature Consistency.} We visualize the rendered language feature map and show that our method provides a consistent intra-instance feature map and a clear boundary, which enhances the segmentation performance.}
    \label{fig:segment}
    \vspace{-0.5cm}
\end{figure}

Previous methods learn semantic features per Gaussian or cluster, causing variations across positions and viewpoints. Our method addresses this issue in Stage 1 through instance-level contrastive learning, and further reinforces feature consistency via a per-instance codebook design. As shown in Figure \ref{fig:3d-eval}, on the \textit{Teatime} and \textit{Waldo\_kitchen} scenes, the feature point clouds produced by LangSplat are highly noisy. On \textit{Scene0097\_00}, the door features from LangSplat are difficult to distinguish, and OpenGaussian oversegments the floor. Figure \ref{fig:segment} visualizes the results with rendered 2D language feature maps. Our method yields homogeneous feature maps with well-defined boundaries.

\subsection{Ablation Study}
All ablations are on \textit{Teatime} of LERF-OVS~\cite{kerr2023lerf}.

\begin{figure}[t]
    \centering
    \includegraphics[width=1.0\linewidth]{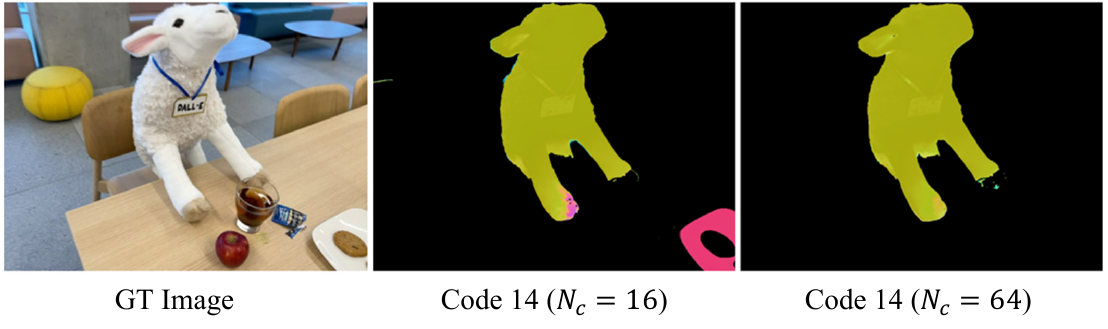}
    \caption{\textbf{Ablation on the Number of Codes.} We visualize an embedding from the codebook as the semantic mask. With codes number $N_c=16$, code 14 represents both the sheep and plate in the \textit{teatime} scene of LERF-OVS. With $N_c=64$, our method clearly isolates the sheep, demonstrating improved instance separation.}
    \label{fig:AblationStudyCodebook}
    \vspace{-0.3cm}
\end{figure}

\noindent \textbf{Ablation on Codebook.}
Table~\ref{table:codebook} shows an ablation on the number of codes. \textit{Teatime} contains around 64 instances; therefore, the best performance is achieved with 64 codes, matching the number of instances and enabling a near one-to-one mapping between embeddings and objects.
Figure~\ref{fig:AblationStudyCodebook} shows that too few codes cause semantic ambiguity. The codebook size matching the expected number of instances achieves the best balance of accuracy and efficiency.

\noindent \textbf{Ablation on Attention Module.}
We also report ablation on the structure of the proposed attention module. In Table \ref{table:ablation}, we show results that a) with lifting MLP Eq.~\ref{eq:lift} only, b) with attention Eq.~\ref{eq:attn} only and set the language codebook as $\mathcal{C}_{lang} \in R^{N_c\times 512}$, c) the attention together with lifting MLP but without residual connection $\mathcal{F}_{lift}\big(\mathcal{A}(Q,K,V)\big)$, e) our full model. We find that our full model achieves the best overall performance. In case b), applying the attention module directly without the MLP leads to high computational cost and convergence difficulties. Comparing cases c) and d), we observe that introducing a residual connection significantly improves training stability.

\noindent \textbf{Ablation on Probability Guidance.}
In Table \ref{table:ablation}, we show results that d) without probability guidance. Our guided attention model is better able to assign distinct embeddings to separate object instances, leading to more accurate and interpretable segmentation. The right column of Figure~\ref{fig:AblationStudyCodebook} visualizes a selected embedding from our proposed codebook, showing that each embedding indeed captures meaningful instance-level semantics. For more ablations and runtime analysis, please refer to \textbf{supp. mat.}.

\begin{table}[t]
\small
\centering
\resizebox{0.3\textwidth}{!}{
\begin{tabular}{lcccc}
\toprule
\textbf{$N_c =$} & {16} & {32} & {64} & {128} \\
\midrule
mIoU $\uparrow$ & 60.22 & 71.65 & \textbf{76.73} & 68.33 \\
mAcc $\uparrow$ & 72.88 & 83.05 & \textbf{88.14} & 83.05 \\
\bottomrule
\end{tabular}
}
\vspace{-0.3cm}
\caption{\textbf{Ablation on Number of Codes.}}
\label{table:codebook}
\vspace{-0.3cm}
\end{table}

\begin{table}[t]
\centering
  \resizebox{0.35\textwidth}{!}{
  \begin{tabular}{ccccccc}
    \toprule
    \# & MLP & Attn & Res. & Prob. & mIoU $\uparrow$ & mAcc $\uparrow$ \\
    \cmidrule(lr){1-1} \cmidrule(lr){2-5} \cmidrule(lr){6-7} 
    a) & \checkmark & \ & \ & \ & 72.60 & 86.44 \\
    b) & \ & \checkmark & \ & \ & 35.48 & 42.37 \\
    c) & \checkmark & \checkmark & \ & \ & 74.25 & 88.13 \\
    d) & \checkmark & \checkmark & \checkmark & \ & 75.02 & 86.44 \\
    e) & \checkmark & \checkmark & \checkmark & \checkmark & \textbf{76.73} & \textbf{88.14} \\
    \bottomrule
  \end{tabular}
  }
  \vspace{-0.3cm}
  \caption{\textbf{Ablation on Attention and Probability Guidance.}}
  \label{table:ablation}
  \vspace{-0.5cm}
\end{table}

\section{Conclusions}
\label{sec:conclusion}
We presented GALA, a framework for open-vocabulary 3D scene understanding using 3D Gaussian Splatting. By combining self-supervised instance-level feature distillation with a cross-attention module and learnable codebooks, GALA produces consistent, view-independent semantic embeddings, supports 2D and 3D open-vocabulary queries, and reduces memory usage. Experiments on real-world datasets demonstrate its effectiveness in generating reliable and efficient 3D and 2D feature representations.

%leave an empty row here, or the template breaks and compresses the lines of the conclusion

{
    \small
    \bibliographystyle{ieeenat_fullname}
    \bibliography{main}

\begin{thebibliography}{47}
\providecommand{\natexlab}[1]{#1}
\providecommand{\url}[1]{\texttt{#1}}
\expandafter\ifx\csname urlstyle\endcsname\relax
  \providecommand{\doi}[1]{doi: #1}\else
  \providecommand{\doi}{doi: \begingroup \urlstyle{rm}\Url}\fi

\bibitem[Anderson et~al.(2018)Anderson, Wu, Teney, Bruce, Johnson, S{\"u}nderhauf, Reid, Gould, and Van Den~Hengel]{anderson2018vision}
Peter Anderson, Qi Wu, Damien Teney, Jake Bruce, Mark Johnson, Niko S{\"u}nderhauf, Ian Reid, Stephen Gould, and Anton Van Den~Hengel.
\newblock Vision-and-language navigation: Interpreting visually-grounded navigation instructions in real environments.
\newblock In \emph{Proceedings of the IEEE conference on computer vision and pattern recognition}, pages 3674--3683, 2018.

\bibitem[Barron et~al.(2021)Barron, Mildenhall, Tancik, Hedman, Martin-Brualla, and Srinivasan]{barron2021mip}
Jonathan~T Barron, Ben Mildenhall, Matthew Tancik, Peter Hedman, Ricardo Martin-Brualla, and Pratul~P Srinivasan.
\newblock Mip-nerf: A multiscale representation for anti-aliasing neural radiance fields.
\newblock In \emph{Proceedings of the IEEE/CVF international conference on computer vision}, pages 5855--5864, 2021.

\bibitem[Caesar et~al.(2020)Caesar, Bankiti, Lang, et~al.]{caesar2020nuscenes}
Holger Caesar, Varun Bankiti, Alex~H. Lang, et~al.
\newblock nuscenes: A multimodal dataset for autonomous driving.
\newblock \emph{Proceedings of the IEEE/CVF Conference on Computer Vision and Pattern Recognition (CVPR)}, 2020.

\bibitem[Cen et~al.(2023)Cen, Zhou, Fang, Shen, Xie, Jiang, Zhang, Tian, et~al.]{cen2023segment}
Jiazhong Cen, Zanwei Zhou, Jiemin Fang, Wei Shen, Lingxi Xie, Dongsheng Jiang, Xiaopeng Zhang, Qi Tian, et~al.
\newblock Segment anything in 3d with nerfs.
\newblock \emph{Advances in Neural Information Processing Systems}, 36:\penalty0 25971--25990, 2023.

\bibitem[Chen et~al.(2023)Chen, Zhao, et~al.]{sceneAR2023}
Jiaqi Chen, Ruoxi Zhao, et~al.
\newblock Scenear: Learning to reconstruct 3d indoor scenes for augmented reality.
\newblock In \emph{Proceedings of the IEEE/CVF Conference on Computer Vision and Pattern Recognition (CVPR)}, 2023.

\bibitem[Cheng et~al.(2024)Cheng, Zaech, Van~Gool, and Paudel]{cheng2024occam}
Jiahuan Cheng, Jan-Nico Zaech, Luc Van~Gool, and Danda~Pani Paudel.
\newblock Occam's lgs: An efficient approach for language gaussian splatting.
\newblock \emph{arXiv preprint arXiv:2412.01807}, 2024.

\bibitem[Dai et~al.(2017)Dai, Chang, Savva, Halber, Funkhouser, and Nie{\ss}ner]{dai2017scannet}
Angela Dai, Angel~X Chang, Manolis Savva, Maciej Halber, Thomas Funkhouser, and Matthias Nie{\ss}ner.
\newblock Scannet: Richly-annotated 3d reconstructions of indoor scenes.
\newblock In \emph{Proceedings of the IEEE conference on computer vision and pattern recognition}, pages 5828--5839, 2017.

\bibitem[Geiger et~al.(2012)Geiger, Lenz, and Urtasun]{geiger2012kitti}
Andreas Geiger, Philip Lenz, and Raquel Urtasun.
\newblock Are we ready for autonomous driving? the kitti vision benchmark suite.
\newblock \emph{Proceedings of the IEEE Conference on Computer Vision and Pattern Recognition (CVPR)}, 2012.

\bibitem[Huang et~al.(2024)Huang, Yu, Chen, Geiger, and Gao]{huang20242d}
Binbin Huang, Zehao Yu, Anpei Chen, Andreas Geiger, and Shenghua Gao.
\newblock 2d gaussian splatting for geometrically accurate radiance fields.
\newblock In \emph{ACM SIGGRAPH 2024 conference papers}, pages 1--11, 2024.

\bibitem[Huang et~al.(2022)Huang, Mees, Zeng, and Burgard]{huang2022visual}
Chenguang Huang, Oier Mees, Andy Zeng, and Wolfram Burgard.
\newblock Visual language maps for robot navigation.
\newblock \emph{arXiv preprint arXiv:2210.05714}, 2022.

\bibitem[Hughes et~al.(2024)Hughes, Chang, Hu, Talak, Abdulhai, Strader, and Carlone]{hughes2024foundations}
Nathan Hughes, Yun Chang, Siyi Hu, Rajat Talak, Rumaia Abdulhai, Jared Strader, and Luca Carlone.
\newblock Foundations of spatial perception for robotics: Hierarchical representations and real-time systems.
\newblock \emph{The International Journal of Robotics Research}, 43\penalty0 (10):\penalty0 1457--1505, 2024.

\bibitem[Jun-Seong et~al.(2025)Jun-Seong, Kim, Yu-Ji, Wang, Choe, and Oh]{jun2025dr}
Kim Jun-Seong, GeonU Kim, Kim Yu-Ji, Yu-Chiang~Frank Wang, Jaesung Choe, and Tae-Hyun Oh.
\newblock Dr. splat: Directly referring 3d gaussian splatting via direct language embedding registration.
\newblock In \emph{Proceedings of the Computer Vision and Pattern Recognition Conference}, pages 14137--14146, 2025.

\bibitem[Kerbl et~al.(2023)Kerbl, Kopanas, Leimk{\"u}hler, and Drettakis]{kerbl20233d}
Bernhard Kerbl, Georgios Kopanas, Thomas Leimk{\"u}hler, and George Drettakis.
\newblock 3d gaussian splatting for real-time radiance field rendering.
\newblock \emph{ACM Trans. Graph.}, 42\penalty0 (4):\penalty0 139--1, 2023.

\bibitem[Kerr et~al.(2023)Kerr, Kim, Goldberg, Kanazawa, and Tancik]{kerr2023lerf}
Justin Kerr, Chung~Min Kim, Ken Goldberg, Angjoo Kanazawa, and Matthew Tancik.
\newblock Lerf: Language embedded radiance fields.
\newblock In \emph{Proceedings of the IEEE/CVF international conference on computer vision}, pages 19729--19739, 2023.

\bibitem[Kim et~al.(2024)Kim, Wu, Kerr, Goldberg, Tancik, and Kanazawa]{kim2024garfield}
Chung~Min Kim, Mingxuan Wu, Justin Kerr, Ken Goldberg, Matthew Tancik, and Angjoo Kanazawa.
\newblock Garfield: Group anything with radiance fields.
\newblock In \emph{Proceedings of the IEEE/CVF Conference on Computer Vision and Pattern Recognition}, pages 21530--21539, 2024.

\bibitem[Kirillov et~al.(2023)Kirillov, Mintun, Ravi, Mao, Rolland, Gustafson, Xiao, Whitehead, Berg, Lo, et~al.]{kirillov2023segment}
Alexander Kirillov, Eric Mintun, Nikhila Ravi, Hanzi Mao, Chloe Rolland, Laura Gustafson, Tete Xiao, Spencer Whitehead, Alexander~C Berg, Wan-Yen Lo, et~al.
\newblock Segment anything.
\newblock In \emph{Proceedings of the IEEE/CVF international conference on computer vision}, pages 4015--4026, 2023.

\bibitem[Lei et~al.(2025)Lei, Wang, Zhou, and Li]{lei2025gaussnav}
Xiaohan Lei, Min Wang, Wengang Zhou, and Houqiang Li.
\newblock Gaussnav: Gaussian splatting for visual navigation.
\newblock \emph{IEEE Transactions on Pattern Analysis and Machine Intelligence}, 2025.

\bibitem[Li et~al.(2025)Li, Wu, Meng, Gao, Zhang, Wang, and Zhang]{li2025instancegaussian}
Haijie Li, Yanmin Wu, Jiarui Meng, Qiankun Gao, Zhiyao Zhang, Ronggang Wang, and Jian Zhang.
\newblock Instancegaussian: Appearance-semantic joint gaussian representation for 3d instance-level perception.
\newblock In \emph{Proceedings of the Computer Vision and Pattern Recognition Conference}, pages 14078--14088, 2025.

\bibitem[Li et~al.(2024{\natexlab{a}})Li, Niemeyer, Navab, and Tombari]{li2024dns}
Kunyi Li, Michael Niemeyer, Nassir Navab, and Federico Tombari.
\newblock Dns-slam: Dense neural semantic-informed slam.
\newblock In \emph{2024 IEEE/RSJ International Conference on Intelligent Robots and Systems (IROS)}, pages 7839--7846. IEEE, 2024{\natexlab{a}}.

\bibitem[Li et~al.(2024{\natexlab{b}})Li, Lyu, Di, Zhai, Lee, and Tombari]{li2024geogaussian}
Yanyan Li, Chenyu Lyu, Yan Di, Guangyao Zhai, Gim~Hee Lee, and Federico Tombari.
\newblock Geogaussian: Geometry-aware gaussian splatting for scene rendering.
\newblock In \emph{European conference on computer vision}, pages 441--457. Springer, 2024{\natexlab{b}}.

\bibitem[Liang et~al.(2024)Liang, Wang, Li, Niemeyer, Gasperini, Navab, and Tombari]{liang2024supergseg}
Siyun Liang, Sen Wang, Kunyi Li, Michael Niemeyer, Stefano Gasperini, Nassir Navab, and Federico Tombari.
\newblock Supergseg: Open-vocabulary 3d segmentation with structured super-gaussians.
\newblock \emph{arXiv preprint arXiv:2412.10231}, 2024.

\bibitem[Liu et~al.(2024)Liu, Zeng, Ren, Li, Zhang, Yang, Jiang, Li, Yang, Su, et~al.]{liu2024grounding}
Shilong Liu, Zhaoyang Zeng, Tianhe Ren, Feng Li, Hao Zhang, Jie Yang, Qing Jiang, Chunyuan Li, Jianwei Yang, Hang Su, et~al.
\newblock Grounding dino: Marrying dino with grounded pre-training for open-set object detection.
\newblock In \emph{European conference on computer vision}, pages 38--55. Springer, 2024.

\bibitem[Lu et~al.(2024)Lu, Yu, Xu, Xiangli, Wang, Lin, and Dai]{lu2024scaffold}
Tao Lu, Mulin Yu, Linning Xu, Yuanbo Xiangli, Limin Wang, Dahua Lin, and Bo Dai.
\newblock Scaffold-gs: Structured 3d gaussians for view-adaptive rendering.
\newblock In \emph{Proceedings of the IEEE/CVF Conference on Computer Vision and Pattern Recognition}, pages 20654--20664, 2024.

\bibitem[Maggio et~al.(2024)Maggio, Chang, Hughes, Trang, Griffith, Dougherty, Cristofalo, Schmid, and Carlone]{maggio2024clio}
Dominic Maggio, Yun Chang, Nathan Hughes, Matthew Trang, Dan Griffith, Carlyn Dougherty, Eric Cristofalo, Lukas Schmid, and Luca Carlone.
\newblock Clio: Real-time task-driven open-set 3d scene graphs.
\newblock \emph{IEEE Robotics and Automation Letters}, 2024.

\bibitem[Mildenhall et~al.(2021)Mildenhall, Srinivasan, Tancik, Barron, Ramamoorthi, and Ng]{mildenhall2021nerf}
Ben Mildenhall, Pratul~P Srinivasan, Matthew Tancik, Jonathan~T Barron, Ravi Ramamoorthi, and Ren Ng.
\newblock Nerf: Representing scenes as neural radiance fields for view synthesis.
\newblock \emph{Communications of the ACM}, 65\penalty0 (1):\penalty0 99--106, 2021.

\bibitem[M{\"u}ller et~al.(2022)M{\"u}ller, Evans, Schied, and Keller]{muller2022instant}
Thomas M{\"u}ller, Alex Evans, Christoph Schied, and Alexander Keller.
\newblock Instant neural graphics primitives with a multiresolution hash encoding.
\newblock \emph{ACM transactions on graphics (TOG)}, 41\penalty0 (4):\penalty0 1--15, 2022.

\bibitem[Niemeyer et~al.()Niemeyer, Manhardt, Rakotosaona, Oechsle, Duckworth, Gosula, Tateno, Bates, Kaeser, and Tombari]{niemeyer2025radsplat}
Michael Niemeyer, Fabian Manhardt, Marie-Julie Rakotosaona, Michael Oechsle, Daniel Duckworth, Rama Gosula, Keisuke Tateno, John Bates, Dominik Kaeser, and Federico Tombari.
\newblock Radsplat: Radiance field-informed gaussian splatting for robust real-time rendering with 900+ fps.
\newblock In \emph{International Conference on 3D Vision 2025}.

\bibitem[Oquab et~al.(2024)Oquab, Darcet, Moutakanni, Vo, Szafraniec, Khalidov, Fernandez, Haziza, Massa, El-Nouby, et~al.]{oquab2024dinov2}
Maxime Oquab, Timoth{\'e}e Darcet, Th{\'e}o Moutakanni, Huy Vo, Marc Szafraniec, Vasil Khalidov, Pierre Fernandez, Daniel Haziza, Francisco Massa, Alaaeldin El-Nouby, et~al.
\newblock Dinov2: Learning robust visual features without supervision.
\newblock \emph{Transactions on Machine Learning Research Journal}, 2024.

\bibitem[Peng et~al.(2023)Peng, Genova, Jiang, Tagliasacchi, Pollefeys, Funkhouser, et~al.]{peng2023openscene}
Songyou Peng, Kyle Genova, Chiyu Jiang, Andrea Tagliasacchi, Marc Pollefeys, Thomas Funkhouser, et~al.
\newblock Openscene: 3d scene understanding with open vocabularies.
\newblock In \emph{Proceedings of the IEEE/CVF conference on computer vision and pattern recognition}, pages 815--824, 2023.

\bibitem[Peng et~al.(2024)Peng, Wang, Liu, Wen, Dong, and Yang]{peng2024gags}
Yuning Peng, Haiping Wang, Yuan Liu, Chenglu Wen, Zhen Dong, and Bisheng Yang.
\newblock Gags: Granularity-aware feature distillation for language gaussian splatting.
\newblock \emph{arXiv preprint arXiv:2412.13654}, 2024.

\bibitem[Qin et~al.(2024)Qin, Li, Zhou, Wang, and Pfister]{qin2024langsplat}
Minghan Qin, Wanhua Li, Jiawei Zhou, Haoqian Wang, and Hanspeter Pfister.
\newblock Langsplat: 3d language gaussian splatting.
\newblock In \emph{Proceedings of the IEEE/CVF Conference on Computer Vision and Pattern Recognition}, pages 20051--20060, 2024.

\bibitem[Qu et~al.(2024)Qu, Dai, Li, Lin, Cao, Zhang, and Ji]{qu2024goi}
Yansong Qu, Shaohui Dai, Xinyang Li, Jianghang Lin, Liujuan Cao, Shengchuan Zhang, and Rongrong Ji.
\newblock Goi: Find 3d gaussians of interest with an optimizable open-vocabulary semantic-space hyperplane.
\newblock In \emph{Proceedings of the 32nd ACM international conference on multimedia}, pages 5328--5337, 2024.

\bibitem[Radford et~al.(2021)Radford, Kim, Hallacy, Ramesh, Goh, Agarwal, Sastry, Askell, Mishkin, Clark, et~al.]{radford2021learning}
Alec Radford, Jong~Wook Kim, Chris Hallacy, Aditya Ramesh, Gabriel Goh, Sandhini Agarwal, Girish Sastry, Amanda Askell, Pamela Mishkin, Jack Clark, et~al.
\newblock Learning transferable visual models from natural language supervision.
\newblock In \emph{International conference on machine learning}, pages 8748--8763. PmLR, 2021.

\bibitem[Rauschnabel et~al.(2022)Rauschnabel, Felix, Hinsch, Shahab, and Alt]{rauschnabel2022xr}
Philipp~A Rauschnabel, Reto Felix, Chris Hinsch, Hamza Shahab, and Florian Alt.
\newblock What is xr? towards a framework for augmented and virtual reality.
\newblock \emph{Computers in human behavior}, 133:\penalty0 107289, 2022.

\bibitem[Ren et~al.(2024)Ren, Liu, Zeng, Lin, Li, Cao, Chen, Huang, Chen, Yan, et~al.]{ren2024grounded}
Tianhe Ren, Shilong Liu, Ailing Zeng, Jing Lin, Kunchang Li, He Cao, Jiayu Chen, Xinyu Huang, Yukang Chen, Feng Yan, et~al.
\newblock Grounded sam: Assembling open-world models for diverse visual tasks.
\newblock \emph{arXiv preprint arXiv:2401.14159}, 2024.

\bibitem[Shen et~al.(2024)Shen, Fu, Chen, Zhang, Li, Sun, Wu, Lin, and Ji]{shen2024aligning}
Yunhang Shen, Chaoyou Fu, Peixian Chen, Mengdan Zhang, Ke Li, Xing Sun, Yunsheng Wu, Shaohui Lin, and Rongrong Ji.
\newblock Aligning and prompting everything all at once for universal visual perception.
\newblock In \emph{Proceedings of the IEEE/CVF Conference on Computer Vision and Pattern Recognition}, pages 13193--13203, 2024.

\bibitem[Shi et~al.(2024)Shi, Wang, Duan, and Guan]{shi2024language}
Jin-Chuan Shi, Miao Wang, Hao-Bin Duan, and Shao-Hua Guan.
\newblock Language embedded 3d gaussians for open-vocabulary scene understanding.
\newblock In \emph{Proceedings of the IEEE/CVF Conference on Computer Vision and Pattern Recognition}, pages 5333--5343, 2024.

\bibitem[Tian et~al.(2025)Tian, Li, Ma, Huang, Zheng, Yin, Li, Lu, and Jia]{tian2025ccl}
Lei Tian, Xiaomin Li, Liqian Ma, Hefei Huang, Zirui Zheng, Hao Yin, Taiqing Li, Huchuan Lu, and Xu Jia.
\newblock Ccl-lgs: Contrastive codebook learning for 3d language gaussian splatting.
\newblock \emph{arXiv preprint arXiv:2505.20469}, 2025.

\bibitem[Vaswani et~al.(2017)Vaswani, Shazeer, Parmar, Uszkoreit, Jones, Gomez, Kaiser, and Polosukhin]{vaswani2017attention}
Ashish Vaswani, Noam Shazeer, Niki Parmar, Jakob Uszkoreit, Llion Jones, Aidan~N Gomez, {\L}ukasz Kaiser, and Illia Polosukhin.
\newblock Attention is all you need.
\newblock \emph{Advances in neural information processing systems}, 30, 2017.

\bibitem[Wang et~al.(2024{\natexlab{a}})Wang, Zhu, Huang, Chen, Zhu, and Lu]{wang2024drivedreamer}
Xiaofeng Wang, Zheng Zhu, Guan Huang, Xinze Chen, Jiagang Zhu, and Jiwen Lu.
\newblock Drivedreamer: Towards real-world-drive world models for autonomous driving.
\newblock In \emph{European conference on computer vision}, pages 55--72. Springer, 2024{\natexlab{a}}.

\bibitem[Wang et~al.(2024{\natexlab{b}})Wang, He, Fan, Li, Chen, and Zhang]{wang2024driving}
Yuqi Wang, Jiawei He, Lue Fan, Hongxin Li, Yuntao Chen, and Zhaoxiang Zhang.
\newblock Driving into the future: Multiview visual forecasting and planning with world model for autonomous driving.
\newblock In \emph{Proceedings of the IEEE/CVF Conference on Computer Vision and Pattern Recognition}, pages 14749--14759, 2024{\natexlab{b}}.

\bibitem[Wu et~al.(2024)Wu, Meng, Li, Wu, Shi, Cheng, Zhao, Feng, Ding, Wang, et~al.]{wu2024opengaussian}
Yanmin Wu, Jiarui Meng, Haijie Li, Chenming Wu, Yahao Shi, Xinhua Cheng, Chen Zhao, Haocheng Feng, Errui Ding, Jingdong Wang, et~al.
\newblock Opengaussian: Towards point-level 3d gaussian-based open vocabulary understanding.
\newblock \emph{Advances in Neural Information Processing Systems}, 37:\penalty0 19114--19138, 2024.

\bibitem[Xu et~al.(2022)Xu, Xu, Philip, Bi, Shu, Sunkavalli, and Neumann]{xu2022point}
Qiangeng Xu, Zexiang Xu, Julien Philip, Sai Bi, Zhixin Shu, Kalyan Sunkavalli, and Ulrich Neumann.
\newblock Point-nerf: Point-based neural radiance fields.
\newblock In \emph{Proceedings of the IEEE/CVF conference on computer vision and pattern recognition}, pages 5438--5448, 2022.

\bibitem[Ye et~al.(2024)Ye, Danelljan, Yu, and Ke]{ye2024gaussian}
Mingqiao Ye, Martin Danelljan, Fisher Yu, and Lei Ke.
\newblock Gaussian grouping: Segment and edit anything in 3d scenes.
\newblock In \emph{European conference on computer vision}, pages 162--179. Springer, 2024.

\bibitem[Ying et~al.(2024)Ying, Yin, Zhang, Wang, Yu, Huang, and Fang]{ying2024omniseg3d}
Haiyang Ying, Yixuan Yin, Jinzhi Zhang, Fan Wang, Tao Yu, Ruqi Huang, and Lu Fang.
\newblock Omniseg3d: Omniversal 3d segmentation via hierarchical contrastive learning.
\newblock In \emph{Proceedings of the IEEE/CVF Conference on Computer Vision and Pattern Recognition}, pages 20612--20622, 2024.

\bibitem[Yu et~al.(2024)Yu, Chen, Huang, Sattler, and Geiger]{yu2024mip}
Zehao Yu, Anpei Chen, Binbin Huang, Torsten Sattler, and Andreas Geiger.
\newblock Mip-splatting: Alias-free 3d gaussian splatting.
\newblock In \emph{Proceedings of the IEEE/CVF conference on computer vision and pattern recognition}, pages 19447--19456, 2024.

\bibitem[Zhou et~al.(2024)Zhou, Chang, Jiang, Fan, Zhu, Xu, Chari, You, Wang, and Kadambi]{zhou2024feature}
Shijie Zhou, Haoran Chang, Sicheng Jiang, Zhiwen Fan, Zehao Zhu, Dejia Xu, Pradyumna Chari, Suya You, Zhangyang Wang, and Achuta Kadambi.
\newblock Feature 3dgs: Supercharging 3d gaussian splatting to enable distilled feature fields.
\newblock In \emph{Proceedings of the IEEE/CVF Conference on Computer Vision and Pattern Recognition}, pages 21676--21685, 2024.

\end{thebibliography}
}
% \linenumbers
\clearpage
\setcounter{page}{1}
\maketitlesupplementary

\appendix

\section{Implementation Details}
\subsection{Hyperparameters}

\noindent \textbf{Gaussian Model.}
We adopt Scaffold-GS~\cite{lu2024scaffold} as our appearance model. We use segmentation and geometry features of 16 dimensions each. Two separate two-layer MLPs are employed as the segmentation and geometry decoders in Eq.\ref{eq:sprawn}, both trained with a learning rate of 0.0005. The number of neural Gaussians is set to $K=3$ for both ScanNet-v2~\cite{dai2017scannet} and LERF-OVS~\cite{kerr2023lerf}. We further extend the appearance Gaussian model with an instance feature $\mathbf{m}_i \in \mathbb{R}^{16}$, trained with a learning rate of 0.00005, while the learning rates for other Gaussian attributes follow~\cite{lu2024scaffold}.

\noindent \textbf{Attention and Codebook Module.}
We set the codebook parameters to $N=64$, $d_{\text{ins}}=16$, and $d_{c}=16$, and the learning rate to 0.001. A two-layer MLP is employed as the lifting decoder $\mathcal{F}_{\text{lift}}$, producing an output with dimension $d_{\text{lang}}=512$.

\noindent \textbf{Training.}
By default, we train for 30k iterations in training stage 1 and 15k iterations in training stage 2. During stage 1, we enable the densification and pruning to keep both the appearance and semantic performance.
% densification and pruning are enabled to preserve both appearance and semantic performance.

\noindent \textbf{ScanNet-v2 Dataset.}
Following OpenGaussian~\cite{wu2024opengaussian}, we selected 8 scenes from ScanNet for evaluation. For each scene, we select 1 frame every 20 frames as keyframes for training. 

\noindent \textbf{Preprocess Mask and Language Feature.}
We follow LangSplat~\cite{qin2024langsplat} to preprocess SAM~\cite{kirillov2023segment} masks and CLIP~\cite{radford2021learning} language features from ground-truth RGB images. We use the large level of masks from SAM.

\section{Evaluation}
2D open-vocabulary semantics provide per-pixel classification of objects or regions in the image plane, making them highly effective for tasks such as image segmentation and navigation. However, they are view-dependent, reflecting only what the camera sees, and may yield inconsistent semantics for the same object across viewpoints due to occlusions. In contrast, 3D open-vocabulary semantics assign per-point classifications in 3D space, preserving object size, location, and spatial relationships. They are particularly useful for applications such as robotic manipulation and AR/VR, but are often sparser. To leverage the advantages of both, we report evaluation metrics in both 2D and 3D, where our method achieves strong performance across both domains. In the following, we show how the 2D and 3D evaluation are performed on different datasets.

\subsection{2D Evaluation}
For 2D evaluation, we first render the 2D language feature maps and then compute mIoU and mACC for evaluation. For LERF-OVS~\cite{kerr2023lerf}, we follow the open-vocabulary query protocol of LangSplat~\cite{qin2024langsplat}.

\subsection{3D Evaluation}
\noindent \textbf{LERF-OVS.}
Following LangSplat~\cite{qin2024langsplat}, for 3D evaluation on LERF-OVS we first use the text queries to select the corresponding Gaussians based on their language features, and then rasterize the selected Gaussians into 2D for further evaluation.

\noindent \textbf{ScanNet-v2.}
ScanNet-v2 by default provides semantic point clouds for 3D evaluation. However, during training, the positions and number of points/Gaussians are updated to improve appearance quality. In contrast, OpenGaussian~\cite{wu2024opengaussian} fixes both positions and numbers to facilitate 3D evaluation, which we argue is unfair as it causes a notable drop in appearance performance. Our evaluation is a reproduction of the evaluation protocol proposed in Dr.Splat~\cite{jun2025dr} as their full code is still not accessible. We adopt a shared, volume-aware evaluation protocol that computes per-voxel intersection-over-union (IoU) and accuracy by jointly considering the ScanNet-v2 ground-truth point cloud and the optimized neural Gaussian language features within a common voxel space.

\subsection{Visualization}
To visualize semantic Gaussian results, the high-dimensional language features need to be converted into 3-dimensional color values. We use the autoencoder from LangSplat~\cite{qin2024langsplat} to mapping from CLIP features to RGB.

\section{Runtime Analysis}
We perform single-GPU training with NVIDIA RTX 3090.
Table \ref{tab:training-memory} presents a runtime analysis of the Teatime scene. While our method incurs slightly longer training time due to the Scaffold-GS structure, it achieves significantly faster inference compared to other approaches. The codebook and attention module are extremely lightweight, requiring only \textbf{0.6 MB} of memory. However, our method generates a full-sized language feature map for each view (requires 2 GB) and applies the cosine similarity loss in Eq.~\ref{eq:cos_loss} between the full-sized ground-truth and the 512-dimensional rendered feature map (requires 10 GB), which is memory-intensive. During inference, the cosine similarity loss is not computed, allowing our method to achieve superior runtime efficiency.

Additionally, our training pipeline consists of only two stages, whereas LangSplat and OpenGaussian require three. Regarding memory usage, both our approach and OpenGaussian generate high-dimensional feature point clouds, while LangSplat uses a compressed feature representation. This accounts for the higher memory demand of our method.

\begin{table}[ht]
  \centering
  \resizebox{\columnwidth}{!}{%
    \begin{threeparttable}
      \begin{tabular}{lcccccccc}
       % l
      %  *{4}{S[table-format=1.1]}   % Memory: S1, S2, S3, Total
       % *{4}{S[table-format=2.1]}   % Training: S1, S2, S3, Total
      %  S[table-format=2.1]         % Inference: Inf
     % }
        \toprule
        & \multicolumn{3}{c}{\textbf{Memory (GB)}}
        & \multicolumn{4}{c}{\textbf{Train Time (min)}}
        & \textbf{Inference Time (sec)} \\
        \cmidrule(lr){2-4}
        \cmidrule(lr){5-8}
        % \cmidrule(lr){9-9}
        \textbf{Method}
        & {S1} & {S2} & {S3}
        & {S1} & {S2} & {S3} & {Total}
        &  \\
        \midrule
        LangSplat     
          & 5 & 2 & 6
          & 18 & 7 & 84 & 109
          & 96.00
          \\
        OpenGaussian 
          & 15 & 9 & 13        % memory
          & 43 & 27 & 5 & 75   % training
          & 0.65               % inference
          \\

        Ours          
          &{12} &{14} & \-\-
          &{107} &{109} & \-\- &{216}
          &{0.31}
          \\
        \bottomrule
      \end{tabular}
    \end{threeparttable}
  }% fin resizebox

  \caption{\textbf{Runtime Analysis.} S1–S3 correspond to stage 1, 2 and 3 of training, respectively.}
  \label{tab:training-memory}
  \vspace{-0.2cm}
\end{table}

\section{More Ablation Study}
% \subsection{Ablation on Densification of OpenGaussian}
% As reported in OpenGaussian~\cite{wu2024opengaussian}, Gaussian densification is disabled for ScanNet evaluation. Figure~\ref{fig:supp_opengaussian} visualizes the language feature point cloud with and without densification: with densification, it appears significantly noisier, whereas without densification it is cleaner but much sparser, resulting in a substantial drop in appearance performance. For a fair comparison, we also report OpenGaussian’s performance for both settings in Table~\ref{table:scannet} of the main paper.

%with and without densification. For a fair comparison, we report the performance of OpenGaussian both with and without densification in Table~\ref{table:scannet} of the main paper.
% and we observe that while densification increases the number of Gaussians and improves appearance quality, it negatively impacts segmentation performance in some scenes. 

\subsection{Ablation on Number of Gaussian}
\begin{figure}[t]
    \centering
    \includegraphics[width=1.0\linewidth]{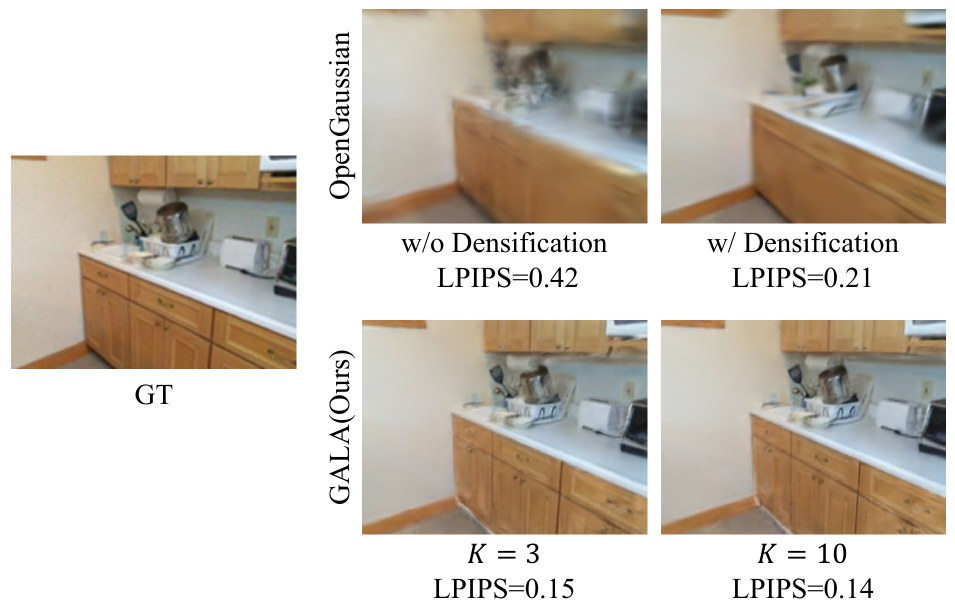}
    \caption{\textbf{Appearance Rendering.} We show the appearance rendering quality on scene0000\_00 of ScanNet-v2 dataset. Compared with OpenGaussian, our method achieve much better appearance rendering.}
    \label{fig:supp_appearance}
\end{figure}

As reported in OpenGaussian~\cite{wu2024opengaussian}, densification is disabled during ScanNet-v2 evaluation, and appearance training is conducted at a very low resolution (160*120). This results in a significant performance drop, as evidenced in Figure~\ref{fig:supp_appearance}, where OpenGaussian produces appearance results that lose many fine details compared to ours.

In contrast, we train with the default resolution as in LangSplat~\cite{qin2024langsplat} (320*240), and Scaffold-GS introduces a parameter $K$ to control the number of spawned Gaussians, making it unnecessary to disable densification entirely. By setting $K=3$ and $K=10$ for the same scene, we can flexibly adjust the number of Gaussians. As shown in Figure~\ref{fig:supp_appearance} and Table~\ref{table:supp_albation}, reducing the number of Gaussians slightly lowers appearance quality, but the degradation is far less severe than in OpenGaussian. More importantly, segmentation performance improves noticeably. This improvement likely arises because semantics carry little or no texture information, so that semantic predictions require fewer Gaussians than appearance modeling. An excessive number of Gaussians may introduce ambiguities that negatively impact segmentation.

% \begin{figure}[t]
%     \centering
%     \includegraphics[width=1.0\linewidth]{figures/supp_opengaussian_scannet.pdf}
%     \caption{\textbf{OpenGaussian~\cite{wu2024opengaussian} with and without Densification.} We visualize the language feature point cloud of OpenGaussian with and without densification. With densification, the semantic point cloud becomes dense but also significantly noisier, whereas without densification, it remains clean but much sparser.}
%     \label{fig:supp_opengaussian}
%     % \vspace{-0.2cm}
% \end{figure}

\begin{table}[t]
\small
\centering
\resizebox{0.35\textwidth}{!}{
\begin{tabular}{@{}lcccc@{}}
\toprule
\textbf{Scenes} & \multicolumn{2}{c}{\textit{Teatime}} & \multicolumn{2}{c}{\textit{Scene0000\_00}} \\
\cmidrule(lr){2-3} \cmidrule(lr){4-5}
\textbf{$K$}    & {3} & {10} & {3} & {10} \\
\midrule
mIoU $\uparrow$ & \textbf{53.27} & 41.68 & \textbf{23.82} & 21.07 \\
mAcc $\uparrow$ & \textbf{84.75} & 71.19 & \textbf{46.83} & 40.07 \\
\bottomrule
\end{tabular}
} 
\caption{\textbf{Ablation on Number of Gaussians.} We report mIoU and mAcc of our proposed method with different number of Gaussians. We report both 3D evaluation on LERF-OVS and ScanNet-v2.}
\label{table:supp_albation}
\vspace{-0.2cm}
\end{table}

\section{More Experiment Results}
\subsection{LERF-OVS}
\begin{figure*}[t]
    \centering
    \includegraphics[width=1.0\linewidth]{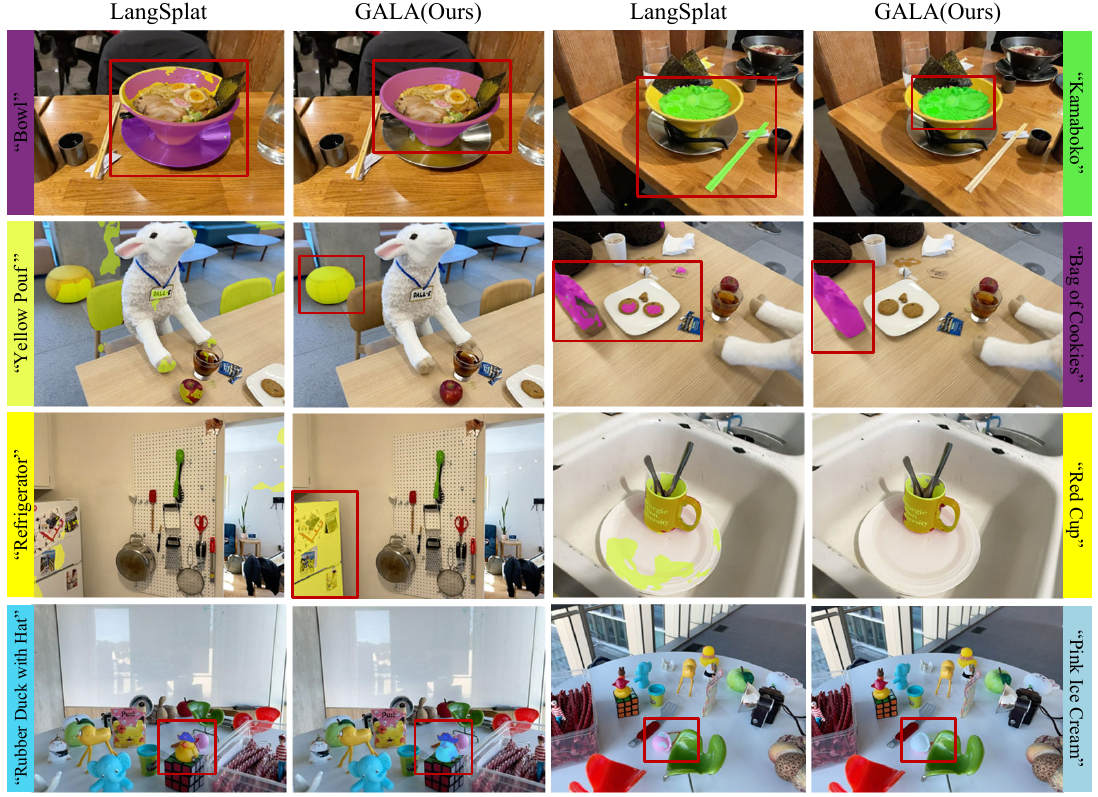}
    \caption{\textbf{More Qualitative Results of 2D Open-Vocabulary Segmentation on LERF-OVS.}}
    \label{fig:supp_2d_lerf}
    % \vspace{-0.2cm}
\end{figure*}

\begin{figure*}[t]
    \centering
    \includegraphics[width=1.0\linewidth]{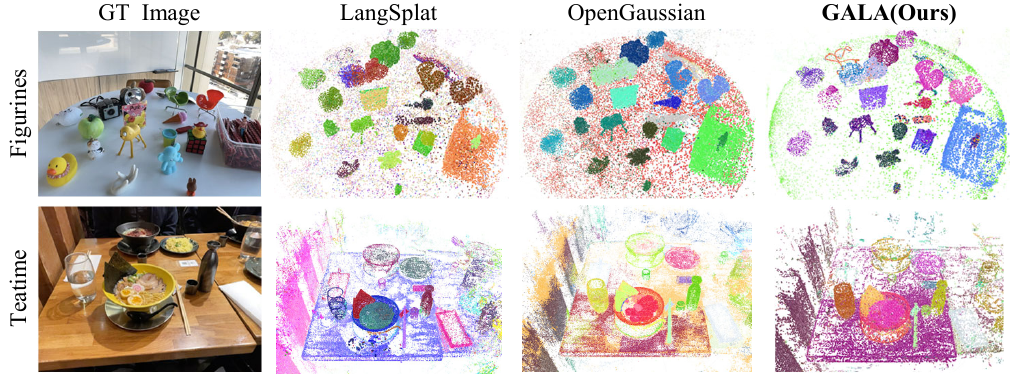}
    \caption{\textbf{More Qualitative Results of 3D Open-Vocabulary Segmentation on LERF-OVS.} We visualize the language feature point cloud by compressing the features into the RGB point cloud. Note that the colors for visualization are consistent only within each method and not method-to-method.}
    \label{fig:supp_3d_lerf}
    % \vspace{-0.2cm}
\end{figure*}

Figure~\ref{fig:supp_2d_lerf}, presents additional 2D qualitative results on the LERF-OVS dataset~\cite{kerr2023lerf}, while Figure~\ref{fig:supp_3d_lerf} shows the corresponding 3D qualitative results.

\subsection{ScanNet-v2}
\begin{figure*}[t]
    \centering
    \includegraphics[width=1.0\linewidth]{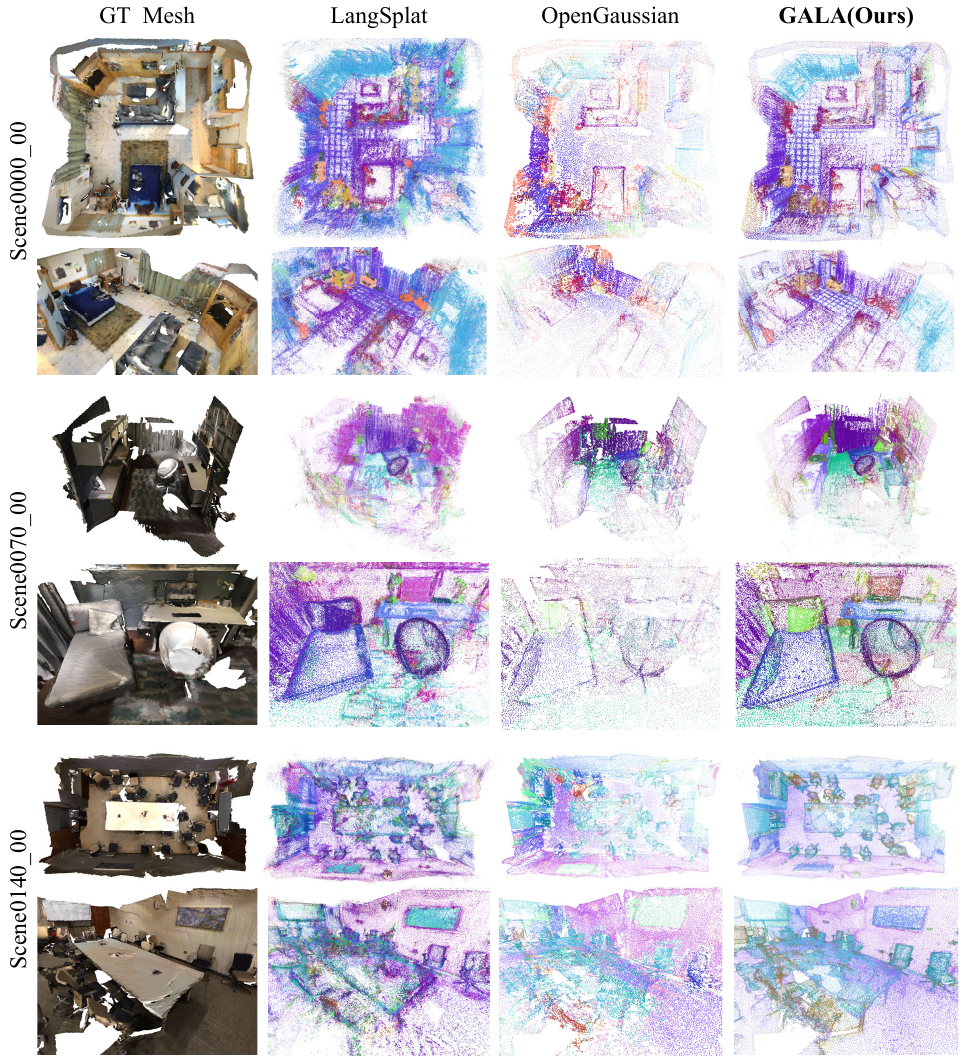}
    \caption{\textbf{More Qualitative Results of 3D Open-Vocabulary Segmentation on ScanNet-v2.} We visualize the language feature point cloud by compressing the features into the RGB point cloud. Note that the colors for visualization are consistent only within each method and not method-to-method. OpenGaussian doesn't enable densification, leads to sparser point cloud.}
    \label{fig:supp_3d_scannet1}
    % \vspace{-0.2cm}
\end{figure*}

\begin{figure*}[t]
    \centering
    \includegraphics[width=1.0\linewidth]{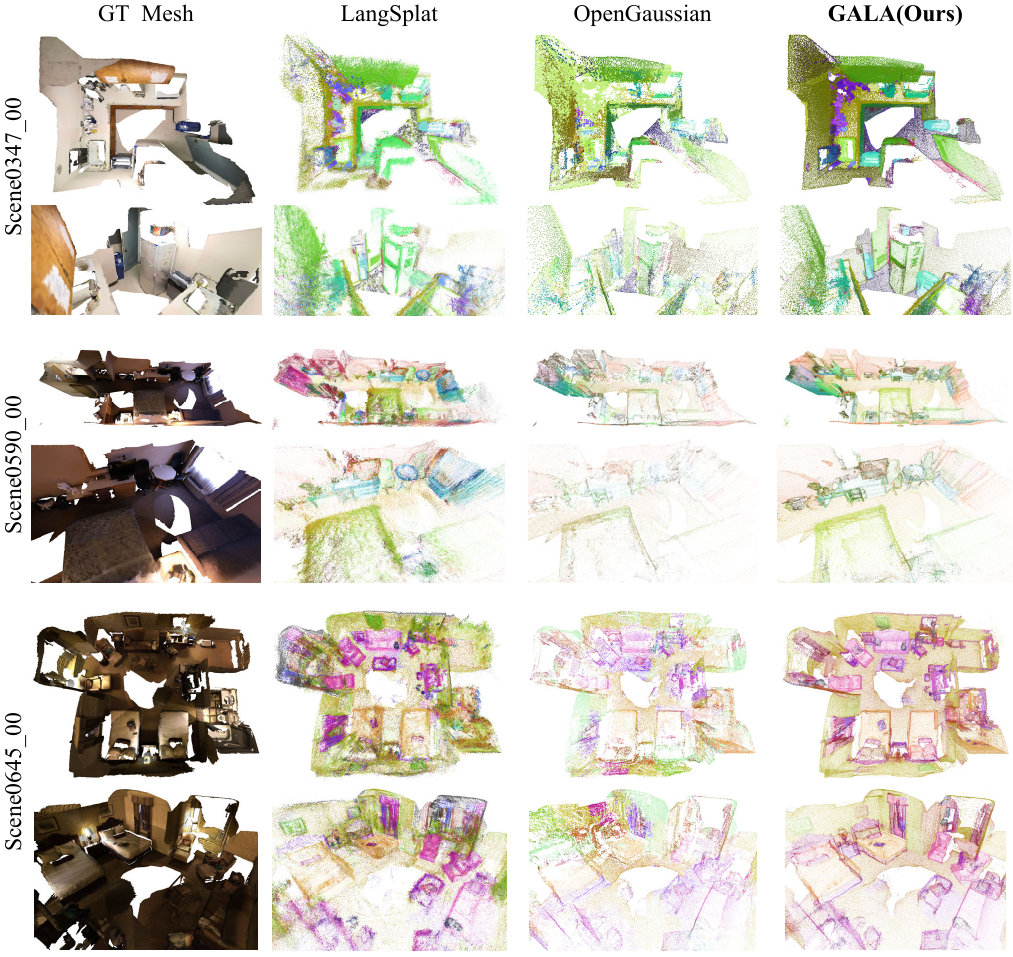}
    \caption{\textbf{More Qualitative Results of 3D Open-Vocabulary Segmentation on ScanNet-v2.} We visualize the language feature point cloud by compressing the features into the RGB point cloud. Note that the colors for visualization are consistent only within each method and not method-to-method. OpenGaussian doesn't enable densification, leads to sparser point cloud.}
    \label{fig:supp_3d_scannet2}
    % \vspace{-0.2cm}
\end{figure*}

In Table~\ref{table:supp_scannet}, we report the per-scene 3D open-vocabulary segmentation and localization results on ScanNet-v2~\cite{dai2017scannet}. In Figures~\ref{fig:supp_3d_scannet1} and Figures~\ref{fig:supp_3d_scannet2}, we show additional qualitative results.

\newpage

\begin{table*}[t]
\centering
\small
\resizebox{1.0\textwidth}{!}{
\begin{tabular}{@{}lccccccccccccccccccc@{}}
\toprule
\textbf{Method} & \multicolumn{2}{c}{\textbf{Mean}} & \multicolumn{2}{c}{\textit{Scene0000\_00}} & \multicolumn{2}{c}{\textit{Scene0062\_00}} & \multicolumn{2}{c}{\textit{Scene0070\_00}} & \multicolumn{2}{c}{\textit{Scene0097\_00}} & \multicolumn{2}{c}{\textit{Scene0140\_00}} & \multicolumn{2}{c}{\textit{Scene0347\_00}} & \multicolumn{2}{c}{\textit{Scene0590\_00}} & \multicolumn{2}{c}{\textit{Scene0645\_00}} \\
\cmidrule(lr){2-3} \cmidrule(lr){4-5} \cmidrule(lr){6-7} \cmidrule(lr){8-9} \cmidrule(lr){10-11} \cmidrule(lr){12-13} \cmidrule(lr){14-15} \cmidrule(lr){16-17} \cmidrule(lr){18-19}
& mIoU $\uparrow$ & mAcc $\uparrow$ & mIoU & mAcc & mIoU & mAcc & mIoU & mAcc & mIoU & mAcc & mIoU & mAcc & mIoU & mAcc & mIoU & mAcc & mIoU & mAcc \\

%\textbf{2D} & & & & & & & & & & & & & & & & & & \\
%\hline
%LangSplat      & 0.0 & 0.0 & 0.0 & 0.0 & 0.0 & 0.0 & 0.0 & 0.0 & 0.0 & 0.0 & 0.0 & 0.0 & 0.0 & 0.0 & 0.0 & 0.0 & 0.0 & 0.0 \\
%OpenGaussian  & 0.0 & 0.0 & 0.0 & 0.0 & 0.0 & 0.0 & 0.0 & 0.0 & 0.0 & 0.0 & 0.0 & 0.0 & 0.0 & 0.0 & 0.0 & 0.0 & 0.0 & 0.0 \\
%\textbf{Ours} & 0.0 & 0.0 & 0.0 & 0.0 & 0.0 & 0.0 & 0.0 & 0.0 & 0.0 & 0.0 & 0.0 & 0.0 & 0.0 & 0.0 & 0.0 & 0.0 & 0.0 & 0.0 \\
%\hline
% \textbf{3D} & & & & & & & & & & & & & & & & & & \\
% \hline
\midrule
%\multicolumn{19}{c}{\textit{Number of Classes: 19}} \\
\multicolumn{2}{l}{\textit{Number of Classes: 19}} & & & & & & & & & & & & & & & & & \\
\midrule
LanSplat~\cite{qin2024langsplat}        & 2.94 & 11.63 & 2.72 & 7.88 & 2.89 & 11.92 & 1.47 & 8.31 & 6.67 & 14.38 & 2.88 & 16.59 & 2.78 & 16.36 & 1.42 & 8.92 & 2.74 & 8.76 \\
OpenGaussian~\cite{wu2024opengaussian}  & 15.47 & 26.04 & 16.12 & 27.67 & 18.02 & 27.42 & \textbf{19.01} & \textbf{31.64} & 9.01 & 13.61 & 15.39 & 28.49 & 23.22 & 35.93 & 11.84 & 18.56 & 11.22 & 25.06 \\
% OpenGaussian*~\cite{wu2024opengaussian} &  16.98 &  25.93 &  8.51 &  15.31 &  17.21 &  30.22 & 12.71 &  19.89 &  14.06 &  16.94 &  \textbf{18.11} &  \textbf{29.06} &  41.04 &  49.83 &  11.08 &  24.26 &  13.14 &  21.97 \\
%\textbf{OursOffset10} & \textbf{21.09} & \textbf{36.85} & \textbf{21.07} & \textbf{40.07} & \textbf{26.73} & \textbf{52.28} &  12.06 & 22.64 & \textbf{15.76} & \textbf{38.67} &  17.87 &  29.05 & \textbf{43.58} & \textbf{49.52} & \textbf{13.71} & \textbf{27.91} & \textbf{17.94} & \textbf{34.70} \\
\textbf{Ours} & \textbf{21.54} & \textbf{37.47} & \textbf{19.04} & \textbf{40.67} & \textbf{20.14} & \textbf{41.01} & 18.54 & 28.88 & \textbf{18.01} & \textbf{38.74} & \textbf{25.55} & \textbf{38.13} & \textbf{43.11} & \textbf{50.13} & \textbf{11.55} & \textbf{29.61} & \textbf{16.40} & \textbf{32.64} \\

\midrule
\multicolumn{2}{l}{\textit{Number of Classes: 15}} & & & & & & & & & & & & & & & & & \\
\midrule
LanSplat~\cite{qin2024langsplat}        & 3.80 & 13.98 & 3.18 & 9.35 & 3.73 & 16.84 & 2.18 & 12.64 & 8.40 & 18.47 & 3.40 & 13.98 & 3.94 & 20.04 & 3.43 & 11.81 & 2.13 & 8.71 \\
OpenGaussian~\cite{wu2024opengaussian}  & 17.42 & 28.82 & 19.18 & 33.77 & \textbf{18.65} & 28.71 & 21.16 & 26.89 & 10.76 & 17.31 & 18.64 & 29.86 &  20.50 &  36.85 & 15.78 & 28.20 & 14.73 & 29.02 \\
% OpenGaussian*~\cite{wu2024opengaussian} & 20.03 & 30.64 & 12.61 & 22.16 & 17.28 & 30.99 & 18.21 & 24.08 & \textbf{12.96} & 17.76 & \textbf{22.68} & 33.87 &  38.42 &  49.07 & \textbf{20.50} & \textbf{38.12} & 17.59 & 29.09 \\
%\textbf{OursOffset10}                           & \textbf{24.69} & \textbf{42.18} & \textbf{20.87} & \textbf{44.24} & \textbf{23.63} & \textbf{58.71} & \textbf{22.60} & 25.95 & 9.14 & \textbf{34.86} & 21.58 & \textbf{34.10} & \textbf{63.02} & \textbf{66.97} & 16.98 & 33.86 & \textbf{19.71} & \textbf{38.82} \\
\textbf{Ours}                           & \textbf{25.20} & \textbf{42.06} & \textbf{23.82} & \textbf{46.83} & 18.23 & \textbf{44.79} & \textbf{29.43} & \textbf{34.22} & \textbf{12.75} & \textbf{37.79} & \textbf{30.50} & \textbf{43.31} & \textbf{50.85} & \textbf{59.61} & \textbf{16.62} & \textbf{34.99} & \textbf{19.41} & \textbf{34.95} \\

\midrule
\multicolumn{2}{l}{\textit{Number of Classes: 10}} & & & & & & & & & & & & & & & & & \\
\midrule
LanSplat~\cite{qin2024langsplat}        & 6.60 & 22.24 & 5.04 & 14.35 & 5.78 & 25.27 & 4.34 & 16.90 & 18.15 & 36.69 & 4.20 & 16.82 & 7.13 & 38.72 & 4.97 & 16.59 & 3.19 & 12.61 \\
OpenGaussian~\cite{wu2024opengaussian}  & 23.46 & 37.73 & 22.70 & 39.32 & 28.34 & 43.10 & 29.09 & 37.55 & 21.72 & 29.88 & 21.91 & 35.11 & 20.34 & 46.22 & \textbf{25.56} & 36.17 & 18.09 & 34.55 \\
% OpenGaussian*~\cite{wu2024opengaussian} & 27.46 & 40.71 & 17.43 & 27.62 & 26.35 & 46.57 & 18.49 & 36.35 & \textbf{26.91} & 35.61 & 26.14 & 39.87 & 51.61 & 67.10 & 32.53 & 43.09 & 20.29 & 29.47 \\
%\textbf{OursOffset10}                  & \textbf{35.03} & \textbf{54.77} & 22.41 & \textbf{51.47} & \textbf{33.19} & \textbf{67.87} & \textbf{37.85} & \textbf{45.44} & 20.43 & \textbf{48.36} & \textbf{31.43} & \textbf{50.33} & \textbf{74.42} & \textbf{75.21} & \textbf{34.98} & \textbf{48.79} & \textbf{25.57} & \textbf{50.73} \\
\textbf{Ours} & \textbf{35.85} & \textbf{57.02} & \textbf{27.61} & \textbf{57.53} & \textbf{30.02} & \textbf{66.91} & \textbf{51.76} & \textbf{67.65} & \textbf{24.65} & \textbf{52.81} & \textbf{45.26} & \textbf{60.62} & \textbf{56.03} & \textbf{56.20} & 25.17 & \textbf{43.74} & \textbf{26.36} & \textbf{50.74} \\

\bottomrule
\end{tabular}
} 
\caption{\textbf{3D Evaluation on ScanNet-v2.} We report per scene mIoU and mAcc on the ScanNet-v2 dataset~\cite{dai2017scannet}, following the evaluation protocol of Dr.Splat~\cite{jun2025dr}.}
\label{table:supp_scannet}
\end{table*}

\end{document}